\documentclass[11pt]{article}

\usepackage[final]{acl}

\usepackage{times}
\usepackage{latexsym}

\usepackage[T1]{fontenc}

\usepackage[utf8]{inputenc}

\usepackage{microtype}

\usepackage{inconsolata}

\usepackage{graphicx}

\title{Instructions for *ACL Proceedings}
\usepackage{booktabs}       
\usepackage{amsfonts}       

\usepackage{graphicx}
\usepackage{amsmath}  
\usepackage{multirow} 
\usepackage[normalem]{ulem}  
\usepackage{arydshln}  
\usepackage[table]{xcolor} 

\usepackage{graphicx} 

\usepackage{enumitem} 

\usepackage{algorithm} 
\usepackage{algorithmic}  

\newcommand{\ours}{$\pi$-Play }
\newcommand{\blackours}{$\boldsymbol{\pi}$-Play }
\newcommand{\blackoursN}{$\boldsymbol{\pi}$-Play}

\usepackage{xcolor,pifont}

\definecolor{blue1}{rgb}{0.118, 0.251, 0.486}
\definecolor{blue2}{rgb}{0.918, 0.949, 0.996}
\definecolor{lightpurple}{rgb}{0.941, 0.92, 1.000} 
\definecolor{darkpurple3}{rgb}{0.35, 0.35, 0.80}
\definecolor{darkpurple}{rgb}{0.25, 0.25, 0.60}   
\definecolor{normalpurple}{rgb}{0.62, 0.436, 0.706}   
\definecolor{lightorange}{rgb}{1, 0.702, 0.278}
\definecolor{lightorange}{rgb}{1, 0.85, 0.60}

\usepackage[most]{tcolorbox}  
\tcbuselibrary{listings,breakable,skins} 
\lstdefinestyle{boxedcode}{
  basicstyle=\ttfamily\small, 
  columns=fullflexible,
  keepspaces=true,
  breaklines=true,
  upquote=true,
  showstringspaces=false
}

\title{$\pi$-Play: Multi-Agent Self-Play via Privileged Self-Distillation without External Data}

\author{%
  Yaocheng Zhang\thanks{Work primarily done during an internship at Meituan.} \ $^{1,2}$, \ \
  Yuanheng Zhu\thanks{Corresponding author} \ $^{1,4}$,\ \
  Wenyue Chong$^{1,2}$, \ \
  Songjun Tu$^{1,4}$,\\
  \textbf{Qichao Zhang}$^{1,4}$, \ \
  \textbf{Jiajun Chai}$^{3}$, \ \
  \textbf{Xiaohan Wang}$^{3}$, \ \
  \textbf{Wei Lin}$^{3}$, \\
  \textbf{Guojun Yin}$^{3}$, \ \
  \textbf{Dongbin Zhao}$^{1,2,4}$ \\
  $^{1}$Institute of Automation, Chinese Academy of Sciences \\
  $^{2}$School of Advanced Interdisciplinary Sciences, University of Chinese Academy of Sciences \\
  $^{3}$Meituan \quad
  $^{4}$School of Artificial Intelligence, University of Chinese Academy of Sciences \\
  \texttt{\{zhangyaocheng2023,yuanheng.zhu\}@ia.ac.cn} \\
}

\begin{document}

\newtcolorbox{promptbox}[1]{
  colback=blue!5,     
  breakable,                 
  colframe=black!75!black,    
  colbacktitle=black,         
  coltitle=white,             
  title=#1,                   
  boxrule=1pt,                
  arc=1mm,                    
  enhanced,
  attach boxed title to top left={xshift=3mm,yshift=-3mm},
  boxed title style={height=6mm},  
  left=3mm,
  right=3mm,
  top=3.0mm,
  bottom=1.0mm,
  width=0.52\textwidth,
  center,
}

\newtcolorbox{contribution}[1]{
  colframe=black!35,    
  colbacktitle=black,         
  coltitle=white,             
  title=#1,                   
  boxrule=1pt,                
  arc=1mm,                    
  enhanced,
  attach boxed title to top left={xshift=3mm,yshift=-3mm},
  boxed title style={height=6mm},  
  left=3mm,
  right=3mm,
  top=2.0mm,
  bottom=1.0mm,
  width=0.48\textwidth,
  center,
}

\maketitle

\begin{abstract}
Deep search agents have emerged as a promising paradigm for addressing complex information-seeking tasks, but their training remains challenging due to sparse rewards, weak credit assignment, and limited labeled data. Self-play offers a scalable route to reduce data dependence, but conventional self-play optimizes students only through sparse outcome rewards, leading to low learning efficiency. In this work, we observe that self-play naturally produces a question construction path (QCP) during task generation, an intermediate artifact that captures the reverse solution process. This reveals a new source of privileged information: self-play can provide high-quality privileged information for the self-distillation at low cost and at scale, without relying on human feedback or curated privileged information. Leveraging this insight, we propose \textbf{P}rivileged \textbf{I}nformation Self-\textbf{Play} ($\pi$-Play), a novel multi-agent self-evolution framework combining self-play and self-distillation. In $\pi$-Play, an \textit{examiner} generates tasks together with QCPs, and a \textit{teacher} employs QCP as privileged context to densely supervise a \textit{student} via self-distillation. This design transforms sparse-reward self-play into a dense-feedback co-evolution. Extensive experiments show that data-free $\pi$-Play surpasses fully supervised search agents and improves evolutionary efficiency by 2–3$\times$ over conventional self-play. Code is available at \url{https://github.com/zhyaoch/pi-play}.
\end{abstract}

\section{Introduction}
Deep search agents leverage the reasoning capabilities of large language models (LLMs) and external search engines to perform multi-turn retrieval and analysis for complex questions, emerging as a promising paradigm for information acquisition \citep{shao2024deepseekmathpushinglimitsmathematical,deepseekai2025deepseekr1incentivizingreasoningcapability,zheng-etal-2025-deepresearcher}. Recent advances have shown that reinforcement learning (RL) can substantially improve both reasoning and search behaviors, enabling LLM agents to tackle increasingly challenging information-seeking tasks \citep{shao2024deepseekmathpushinglimitsmathematical,deepseekai2025deepseekr1incentivizingreasoningcapability}. However, training strong search agents at scale remains fundamentally bottlenecked by data \citep{yue2026drzeroselfevolvingsearch,jin2025searchr1trainingllmsreason,wu2025webdancer, li2025websailor, gao2025turnsunlockinglonghorizonagentic}. Supervised pipelines rely on labeled data and costly expert trajectories, while outcome-supervised RL often suffers from sparse rewards and poor credit assignment, especially in multi-turn search scenarios \citep{jin2025searchr1trainingllmsreason,zhang2025criticsearchfinegrainedcreditassignment,tu2026dynamic,wang2025stepsearchignitingllmssearch,zyc2025aamas,yue2025promotingefficientreasoningverifiable}.

\begin{figure*}[t]
    \centering
    \includegraphics[width=1\linewidth]{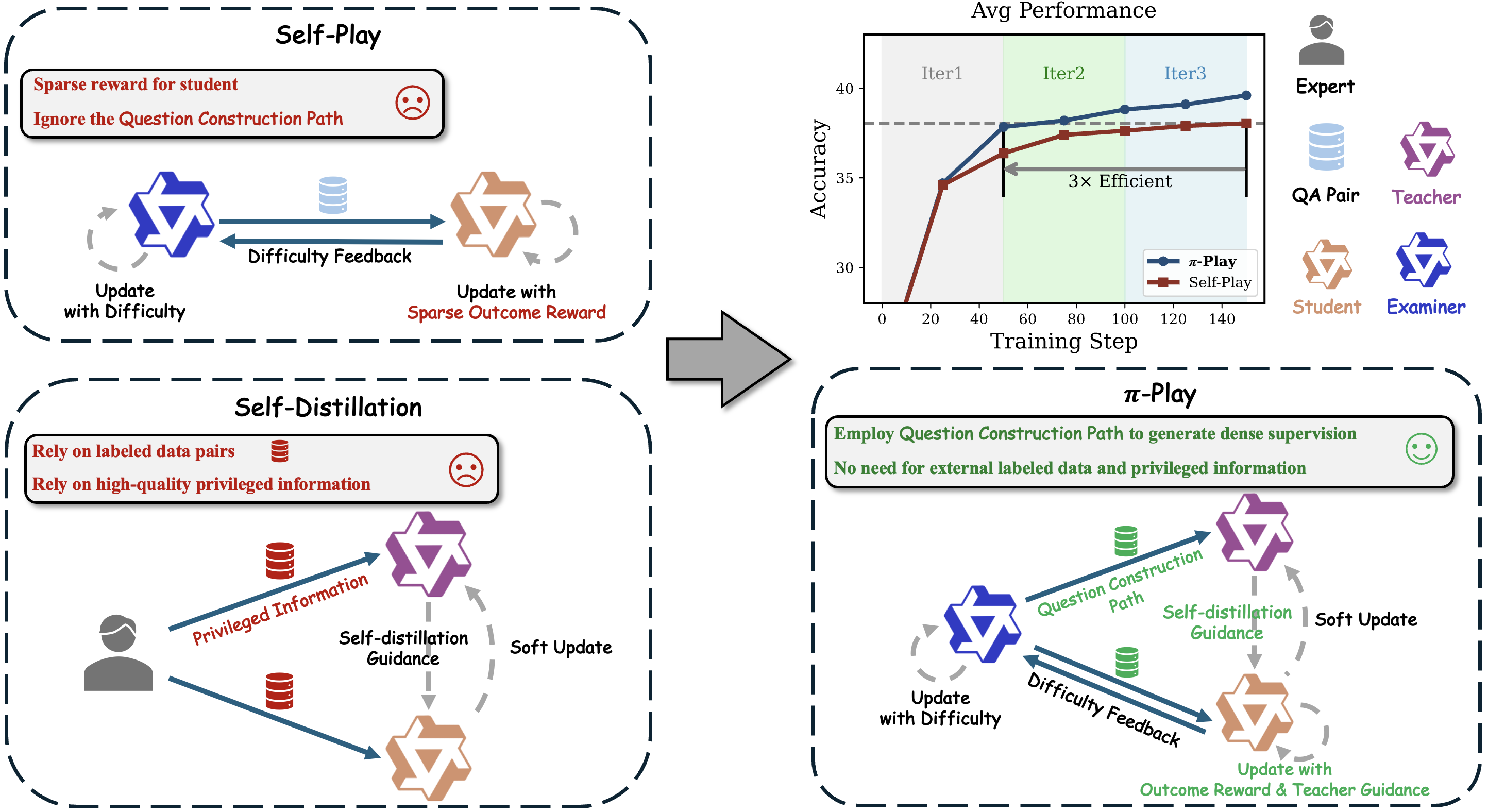}
    \caption{\textbf{Comparison of {$\boldsymbol{\pi}$-Play} with other self-evolution frameworks.} All models (examiner, teacher, and student) in \ours are initialized from the same base LLM and function as search agents. \ours uses alternating optimization to evolve multiple agents in a closed loop. Compared to self-play, it overcomes the sparse-reward of the student and enables the student to be optimized under the joint effect of outcome rewards and teacher guidance.}
    \label{fig:framework_compare}
    \vspace{-10pt}
\end{figure*}

\begin{figure*}[t]
    \centering
    \includegraphics[width=0.97\linewidth]{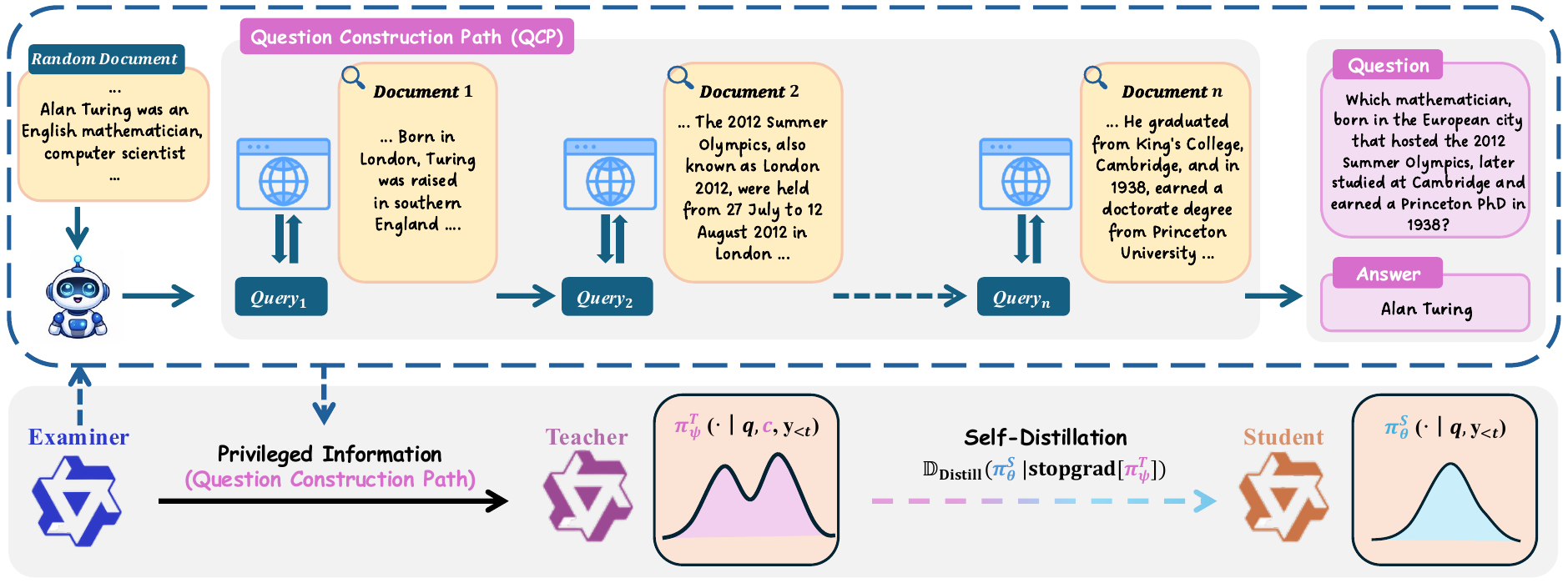}
    \caption{\textbf{Overview of QCP-guided self-distillation in $\pi$-Play.} 
    The examiner is equipped with search tools and interacts with the search engine to obtain factual information, ensuring the correctness of both the synthesized QA pairs and their construction paths $c$. The teacher policy ${\pi_{\psi}^{T}}$ leverages QCP as additional context to provide token-level supervision to the student policy ${\pi_{\theta}^{S}}$ along the student’s rollout $y$, by minimizing the per-token reverse KL divergence $\mathbb{D}_{\mathrm{Distill}}\bigl({\pi_{\theta}^{S}} \,\|\, \text{stopgrad}[{\pi_{\psi}^{T}}]\bigr)$ (Eq.~\ref{Eq:student_distill}).}
    \label{fig:QA_path}
\end{figure*}

A promising direction for alleviating data dependence is self-evolution through \emph{self-play}. In existing self-play frameworks, models of the same scale alternately play the roles of examiner and student: the examiner autonomously constructs training tasks, while the student learns by solving those tasks \citep{yue2026drzeroselfevolvingsearch,huang2026rzeroselfevolvingreasoningllm,lu2025searchselfplaypushingfrontier,openai2021asymmetricselfplayautomaticgoal,chen2024selfplayfinetuningconvertsweak}. This paradigm allows the model to bootstrap its own curriculum without relying on manually collected training data, offering a promising route toward scalable and autonomous improvement. Despite its strengths, self-play still suffers from a critical limitation. The student is typically optimized only through sparse outcome rewards, which makes learning inefficient for multi-turn search tasks \citep{yue2026drzeroselfevolvingsearch}. Notably, self-play produces more than just the final training question-answer (QA) pair $(q, o^{\star})$. 
As shown in Fig.~\ref{fig:QA_path}, self-play also naturally produces a high-quality yet previously overlooked intermediate artifact, the \emph{question construction path} (QCP), denoted by $c$, which captures the multi-turn interaction process by which the examiner constructs the question through iterative search.
Hence, the examiner’s output is more accurately represented as a triplet $(q, c, o^{\star})$, where the QCP records a reverse solution process from the answer back to the question. Rather than being merely an intermediate artifact, the QCP serves as a form of intrinsic privileged information for dense supervision. Unfortunately, existing self-play methods largely ignore this signal, since they cannot directly use it for supervised fine-tuning (SFT) \citep{yue2026drzeroselfevolvingsearch,huang2026rzeroselfevolvingreasoningllm,lu2025searchselfplaypushingfrontier,fu2025srftsinglestagemethodsupervised}, and thus fail to exploit it as a valuable source of dense supervision.

Another line of self-evolution, \emph{self-distillation}, addresses the credit assignment problem by employing high-quality privileged information \citep{2026reinforcementlearningselfdistillation, shenfeld2026selfdistillationenablescontinuallearning,zhao2026selfdistilledreasoneronpolicyselfdistillation,ye2026onpolicycontextdistillationlanguage, penaloza2026privilegedinformationdistillationlanguage}. Unlike on-policy distillation\citep{agarwal2024onpolicy,gu2024minillm,yue2025does,lu2025onpolicydistillation}, which relies on a larger external teacher, self-distillation uses a teacher model of the same scale as the student and augments it with privileged information to provide token-level dense supervision for the student. Common sources of such privileged information include expert demonstrations \citep{penaloza2026privilegedinformationdistillationlanguage,shenfeld2026selfdistillationenablescontinuallearning}, external (human) feedback \citep{2026reinforcementlearningselfdistillation, wang2026openclawrltrainagentsimply}, and prior knowledge \citep{ye2026onpolicycontextdistillationlanguage,sang2026crispcompressedreasoningiterative}. Prior studies have shown that such privileged supervision can significantly enhance learning efficiency. However, obtaining high-quality privileged information is often nontrivial. In several prior works, privileged information is typically constructed with the help of human experts or stronger models \citep{shenfeld2026selfdistillationenablescontinuallearning,zhao2026selfdistilledreasoneronpolicyselfdistillation,ye2026onpolicycontextdistillationlanguage}. Furthermore, self-distillation typically relies on training data consisting of QA pairs during optimization. This dependence on both high-quality privileged information and curated training data makes self-distillation difficult to scale efficiently.

This work is motivated by a key observation: self-play naturally produces intrinsic privileged information that can be exploited for self-distillation. Since QCP records how the question is constructed from factual evidence, it provides privileged context that can help a same-scale teacher to generate more accurate rollouts than a student conditioned only on the question. Based on this insight, we propose \textit{\textbf{P}rivileged \textbf{I}nformation Self-\textbf{Play}} ($\pi$-Play), a multi-agent self-evolution framework in which an examiner first generates training tuples $(q, c, o^{\star})$, and a teacher model conditioned on the construction path $c$ then provides token-level distillation signals to guide the student (Fig.~\ref{fig:QA_path}). Through this mechanism, $\pi$-Play improves search and reasoning capability while transforming sparse-reward self-play into a dense-feedback self-evolution loop. Through multiple efficient training iterations guided by the teacher, \ours surpasses supervised baselines and conventional self-play baselines and exhibits stronger evolutionary efficiency (Fig.\ref{fig:framework_compare}). In summary, our main contributions are as follows:

\begin{enumerate}[leftmargin=4.0mm,label=$\circ$]
        \item We propose {$\pi$-Play}, a novel multi-agent self-evolution framework that combines self-play and self-distillation, in which QCP bridges the examiner, teacher, and student, enabling their efficient co-evolution without external training data.
        \item We reveal that QCP is a new source of high-quality privileged information and that self-play can generate QCPs during task construction at low cost and at scale. This enables QCP-guided self-distillation without relying on human feedback or curated privileged data.
        \item We introduce a teacher role into self-play to transform QCPs into dense token-level supervision for the student through QCP-guided self-distillation. Extensive experiments show that \ours surpasses fully supervised search agents and achieves 2--3$\times$ higher evolutionary efficiency than conventional self-play.
    \end{enumerate}

\section{Methods}
\vspace{-0.3em}

\subsection{Self-Play with Privileged Information}
As shown in Fig.~\ref{fig:framework_compare}, we employ a self-evolution framework based on efficient multi-agent collaboration, in which all models function as search agents capable of leveraging external knowledge. Let $y$ denote a full response (or rollout), and $o(y)$ denote the final answer extracted from $y$; for brevity, we write $o_i := o(y_i)$. The \textit{examiner} ${\pi_\phi^E}$, the \textit{teacher} ${\pi_\psi^T}$, and the \textit{student} ${\pi_\theta^S}$ are each optimized according to their respective objectives:
\begin{promptbox}{Optimization Objective}
\textbf{Examiner} (Sec.~\ref{sec:examiner_optimization}): Learn to generate diverse and challenging questions.
    \begin{equation}
    \small
    {
    \max_{\phi}\mathbb{E}_{\substack{ (q,c,o^{\star})\sim {\pi_\phi^E}(\cdot) , \{ y_{k} \}_{k=1}^n \sim {\pi_{\theta}^S}(\cdot | q) }} [ r^d(o^{\star}, \{ o_k \}_{k=1}^n) ],}
    \end{equation}
\textbf{Teacher} (Sec.~\ref{sec:teacher_optimization}): Incorporate privileged information to generate more accurate rollouts while avoiding an excessive discrepancy between the teacher and the student.
     \begin{equation}
     \small
     {
     \begin{aligned}
    \max_{\psi}&{{\mathbb{E}_{(q,c,o^{\star})\sim {\pi_\phi^E}(\cdot),y \sim {\pi_{\psi}^{T}}(\cdot | q, \textcolor{red}{c})}} [ \mathbb{I}(o=o^{\star})}\\ &-{{\beta \mathbb{D}_{\mathrm{KL}}\bigl( {\pi_{\psi}^{T}}(\cdot \mid q, c)\,\|\, \mathrm{stopgrad}[{\pi_{\theta}^S}(\cdot \mid q)])}}],
    \end{aligned}
    \label{eq:teacher_objective}}
    \end{equation}
\textbf{Student} (Sec.~\ref{sec:student_optimization}): Learns efficiently through a combination of outcome reward and teacher guidance, enhancing its evolutionary efficiency.
    \begin{equation}
    \small
    {
    \begin{aligned}
    \max_{\theta}&{\mathbb{E}_{(q,c,o^{\star})\sim {\pi_\phi^E}(\cdot),y \sim {\pi_{\theta}^S}(\cdot | q)}[ \mathbb{I}(o=o^{\star})} - \\ &{\lambda\textcolor{red}{\mathbb{D}_{\mathrm{Distill}}\bigl({\pi_{\theta}^S}(\cdot \mid q) \,\|\, \mathrm{stopgrad}[{\pi_{\psi}^{T}}(\cdot \mid q, c)])}}],
    \end{aligned}}
    \end{equation}
 
\end{promptbox}
\noindent where $r^d$ denotes the difficulty reward and $\mathbb{I}$ is the indicator function. To generate questions of moderate difficulty (i.e., suitable for the student's current capability), the examiner's reward $r^d$ is defined over the distribution of predicted answers. If all predictions (i.e., $\{ o_k \}_{k=1}^n$) are correct, the question is considered trivial, whereas if none are correct, the question is likely too difficult for the student. To jointly optimize the examiner, teacher, and student, we adopt an alternating optimization loop that couples question generation, teacher guidance, and student improvement into a unified self-evolution process. As the student becomes stronger, it drives the examiner to generate increasingly challenging questions. Meanwhile, the student's updated behavior is softly propagated to the teacher, allowing the teacher to track the student while remaining more stable.
This iterative process establishes a continuously evolving curriculum. 
All three agents are initialized from the same base LLM and rely exclusively on the search tool to access external knowledge. 
Following prior work \citep{yue2026drzeroselfevolvingsearch}, we strictly adhere to a \textit{training data-free setting}, avoiding any demonstrations, questions, or annotated answers  from external sources or human experts.

\subsection{Examiner Training}
\label{sec:examiner_optimization}

To utilize student feedback for training the examiner, we employ a difficulty reward function $r^d$ that encourages both verifiability (the task must be solvable) and difficulty (the task must not be trivial) \citep{yue2026drzeroselfevolvingsearch}. Specifically, we leverage the student's success rate on the generated questions as a proxy for these properties. Let $k$ denote the number of correct solutions out of $n$ sampled attempts. We penalize cases where the student either fails completely ($k = 0$) or succeeds trivially ($k = n$), thereby encouraging the examiner to generate questions of moderate difficulty. The reward is defined as:
\begin{equation}
\small
{\begin{aligned}
r^d(o^{\star}, \{ o_i \}_{i=1}^n) = \mathbb{I}(0 < k < n) \frac{n - k}{n - 1} + r^f,
\end{aligned}
\label{Eq:examiner_reward}}
\end{equation}
where $k = \sum_{i=1}^n \mathbb{I}( o_{i}=o^{\star})$. The difficulty reward is maximized when exactly one solution is correct and decays linearly as the number of correct predictions increases. Here, $r^f$ denotes a format reward that encourages the examiner to interleave reasoning with search during task generation, so that the synthesized questions are both factually grounded and accompanied by an informative construction path. These paths serve as privileged information, enabling the teacher to provide guidance to student. 

Following prior work, we employ hop-grouped relative policy optimization \citep{yue2026drzeroselfevolvingsearch} to train the examiner. Specifically, we estimate advantages by grouping structurally similar questions within a batch. Generated QA pairs are clustered according to their \textit{cross-hop complexity}, measured by the number of hops $h \in \mathcal{H}$. Intuitively, questions with fewer hops are typically simpler, whereas higher-hop questions demand extensive search and multi-turn reasoning. This hop-specific normalization of returns produces low-variance advantage estimates while avoiding the computational cost of sampling multiple candidate questions per prompt. 
\begin{equation}
\small
\begin{aligned}
    \mathcal{J}_{\text{Examiner}}&(\phi) = \mathbb{E}_{\substack{\{ (q_i,c_i,o_i^{\star}) \sim {\pi_\phi^E}(\cdot), \{ y_{i, k} \}_{k=1}^n \sim {\pi_{\theta}^S}(\cdot | q_i) \}_{i=1}^N}} \\
    &\Bigg[ \frac{1}{N} \sum_{h \in \mathcal{H}} \sum_{i \in \mathcal{I}_h} \log {\pi_\phi^E} A_{i,h}  - \beta \mathbb{D}_{\text{KL}}({\pi_\phi^E} \| \pi_{\text{ref}}) \Bigg],
\end{aligned}
\label{Eq:examiner_loss}
\end{equation}
where $N$ denotes the batch size and $\beta$ controls the strength of the KL regularization. The advantage of each generated QA triplet $(q_i, c_i, o_i^{\star})$ is obtained by hop-wise reward normalization:
\begin{equation}
\small
\begin{aligned}
A_{i,h} = \frac{r_i^d - \mathbb{E}_{j \in \mathcal{I}_h}[r_j^d]}{\sqrt{\mathbb{V}\text{ar}_{j \in \mathcal{I}_h}[r_j^d]} + \delta}
\end{aligned}
\label{Eq:examiner_advantage}
\end{equation}
where $\mathcal{I}_h$ is the set of questions in the hop group $h$, and $\delta$ is a small constant for numerical stability.

\subsection{Student Training}
\label{sec:student_optimization}
For student training, we sample training tuples $(q, c, o^{\star})$ from the examiner policy ${\pi_\phi^E}$. The student policy ${\pi_\theta^S}$ then generates candidate rollouts for each question and is optimized using a hybrid training objective. Importantly, the construction path $c$ is only visible to the teacher model, ensuring that the teacher can produce reliable guidance while the student learns to solve the task without direct access to privileged information. Formally, the student objective is defined as follows:
\begin{equation}
\small
\begin{aligned}
&\mathcal{J}_{\text{$\pi$-play}}(\theta) = \mathbb{E}_{(q,c,o^{\star})\sim \pi_\phi^E,\{y_i\}_{i=1}^G \sim {\pi_{\theta}^S}(\cdot | q)} \\ &\Bigg[ 
\underbrace{\frac{1}{G} \sum_{i=1}^G\sum_{t=1}^{|y_i|}\mathcal{L}_{i,t}  - \beta \, \mathbb{D}_{\text{KL}}({\pi_{\theta}^S} \| {\pi_{\text{ref}}})}_{\text{Learning from outcome reward}} - \underbrace{\textcolor{red}{\lambda \mathbb{D}_{\mathrm{Distill}}\bigl({\pi_{\theta}^S} \,\|\, {\pi_{\psi}^{T}})}}_{\text{Teacher guidance}}
\Bigg] 
\end{aligned},
\label{Eq:student_loss}
\end{equation}
\begin{equation}
\small
\mathcal{L}_{i,t}=
\min\!\left(
w_{i,t}A_{i},
\operatorname{clip}\!\left(w_{i,t},
1-\epsilon,1+\epsilon
\right)A_{i}
\right),
\end{equation}
The overall objective consists of two complementary components. The first term corresponds to group relative policy optimization (GRPO), which improves the student policy using outcome rewards derived from answer correctness (i.e., $r^{e}(q, o_i) = \mathbb{I}(o_i = o^{\star})$). The importance weight $w_{i,t}$ and normalized advantage $A_{i}$ are given by:
\begin{equation}
\small
w_{i,t}=\frac{{\pi_{\theta}^S}(y_{i,t}\mid q,y_{i,<t})}{{\pi_{\theta_{\text{old}}}^S}(y_{i,t}\mid q,y_{i,<t})}, \quad A_{i}=\frac{r_i^{e} - {\mathbb{E}_{j\in G}[r_j^{e}]}}{\sqrt{\mathbb{V}\text{ar}_{j\in G}[r_j^{e}]}+ \delta}.
\end{equation}
By computing advantages from group statistics, GRPO reinforces successful rollouts while penalizing failed ones.
The second term is a self-distillation objective, where the student policy is aligned with the teacher policy ${\pi_\psi^T}$ along
student’s rollout. The distillation loss is defined as follows:
\begin{equation}
\small
\begin{aligned}
\mathbb{D}_{\mathrm{Distill}}\bigl({\pi_{\theta}^S}
 \,\|\, {\pi_{\psi}^{T}})=& \frac{1}{|y_i|}\sum_{t=1}^{|y_i|} \mathrm{KL}({\pi_{\theta}^S}(\cdot \mid q, y_{i,<t}) \\ &\| \mathrm{stopgrad}({\pi_{\psi}^{T}}
(\cdot \mid q, \textcolor{red}{c}, y_{i,<t})))
\end{aligned}
\label{Eq:student_distill}
\end{equation}
The full formulation of the distillation loss is provided in Appendix~\ref{app:distillation_loss}. Since the teacher has access to privileged information in the form of the QCP, it can provide reliable token-level guidance. This dense supervision complements sparse outcome rewards through a favorable bias-variance trade-off \citep{SchulmanMLJA15,gu2016q}: outcome rewards are unbiased but high-variance, whereas teacher guidance may introduce modest bias while substantially reducing variance. Their combination enables more efficient credit assignment and faster policy improvement.

\begin{table*}[t]
\centering
\caption{\textbf{The main results of \blackoursN.} \textbf{Bold} values indicate the best result; \uline{underline} values denote the second-best. Through efficient teacher guidance, \ours outperforms both supervised and self-play search agents.}
\resizebox{1\linewidth}{!}{
\setlength{\extrarowheight}{0pt} 
\begin{tabular}{lcccccccc}
\toprule
   & \multicolumn{3}{c}{\textbf{General QA}} & \multicolumn{4}{c}{\textbf{Multi-Hop QA}} &  \\
\cmidrule(lr){2-4} \cmidrule(lr){5-8}

\multirow{2}{*}{}    & \textbf{NQ}    & \textbf{TriviaQA} & \textbf{PopQA} & \textbf{HotpotQA} & \textbf{2WikiMQA} & \textbf{MuSiQue} & \textbf{Bamboogle} & \textbf{Total} \\ 

\cmidrule(l){2-9} & \multicolumn{8}{c}{\textit{\textbf{Qwen3-4B}}} \\ \midrule
ReAct         & 7.2  & 16.6 & 10.3 & 11.2 & 12.7 & 1.6 & 8.8  & 68.4 \\
Search-R1      & 35.2 & 57.3 & 39.5 & 30.7 & 28.9 & 10.5 & \textbf{37.6} & 239.7 \\
SQLM*         & \underline{39.9} & \underline{60.4} & 42.1 & \textbf{33.0} & \underline{30.8} & \underline{10.8} & 29.6 & \underline{246.6} \\
Dr.Zero       & \textbf{40.4} & \textbf{61.2} & \textbf{44.3} & 31.0 & 29.3 & 7.8 & 31.2 & 245.2 \\
\rowcolor{lightpurple}\ours    & 38.1 & 58.4 & \underline{42.3} & \underline{32.3} & \textbf{32.5} & \textbf{11.2} & \underline{35.2} & \textbf{250.0} \\

\cmidrule(l){2-9} & \multicolumn{8}{c}{\textit{\textbf{Qwen3-4B-Instruct}}} \\ \midrule
ReAct         & 18.4 & 45.5 & 28.4 & 22.8 & 18.0 & 6.3 & 27.2 & 166.6 \\
Search-R1      & 39.5 & \underline{62.6} & 41.3 & 34.9 & 24.8 & 12.9 & \textbf{50.0} & 266.0 \\
SQLM*      & 39.0 & 62.2 & \underline{41.9} & 34.1 & 27.3 & 10.2 & 35.2 & 249.9 \\
Dr.Zero       & \underline{41.3} & 61.6 & \textbf{44.0} & \underline{38.4} & \underline{29.8} & \textbf{14.4} & 36.8 & \underline{266.3} \\
\rowcolor{lightpurple}\ours         & \textbf{41.8} & \textbf{63.2} & \textbf{44.0} & \textbf{38.5} & \textbf{32.3} & \underline{13.4} & \underline{44.0} & \textbf{277.2} \\
\cmidrule(l){2-9} & \multicolumn{8}{c}{\textit{\textbf{Qwen3-8B}}} \\ \midrule

ReAct     & 15.9 & 37.3 & 14.9 & 22.3 & 22.5 & 7.4 & 28.8 & 149.1 \\
Search-R1      & 35.7 & 59.9 & 40.0 & 32.0 & 27.8 & 9.8 & 37.6 & 242.8 \\
ToolForge & 30.3 & 16.8 & 35.2& -     &  -    & \textbf{37.8} & \textbf{48.0} & - \\
SQLM*     & 38.9 & 60.5 & 42.7 & 33.1 & 31.7 & 9.9 & 32.0 & 248.8 \\
Dr.Zero       & \underline{42.1} & \underline{62.5} & \underline{45.8} & \underline{36.6} & \textbf{34.8} & \underline{13.1} & \underline{40.0} & \underline{274.9} \\
\rowcolor{lightpurple}\ours   & \textbf{43.0} & \textbf{64.6} & \textbf{47.2} & \textbf{38.9} & \underline{34.2} & 12.4 & \underline{40.0} & \textbf{280.3} \\

\bottomrule

\label{tab:main_result}
\end{tabular}
}
\vspace{-1.5em}
\end{table*}

\subsection{Teacher Updating}
\label{sec:teacher_optimization}
Although Eq.~(\ref{eq:teacher_objective}) defines the ideal teacher objective, directly optimizing it would introduce additional computational overhead. In practice, to provide stable teacher supervision while allowing the teacher to co-evolve with the student, we approximate this objective by updating the teacher parameters as an exponential moving average (EMA) of the student parameters \citep{2026reinforcementlearningselfdistillation, penaloza2026privilegedinformationdistillationlanguage}:
\begin{equation}
\small
\psi \leftarrow (1-\tau)\psi + \tau\theta, \quad \tau \in (0,1),
\label{Eq:teacher_update}
\end{equation}
where $\tau$ controls the teacher update rate. This update keeps the teacher relatively stable while enabling it to gradually track the student's improvements over training.

\begin{algorithm}[t]
\caption{The Evolution Process of \ours}
\normalsize
\begin{algorithmic}[1]
\REQUIRE Examiner model ${\pi_\phi^E}$, Teacher model ${\pi_\psi^T}$, Student model ${\pi_\theta^S}$, Base LLM ${\pi_{\text{ref}}}$

\vspace{1mm}
\STATE Initialize ${\pi_\phi^E}$, ${\pi_\psi^T}$, and ${\pi_\theta^S}$ from Base LLM ${\pi_{\text{ref}}}$
\FOR{iteration $j \leftarrow 1$ to $M$}
    \STATE \textit{\textcolor{gray}{\# Updating Examiner Model}}
    \FOR{step $k \leftarrow 1$ to $W$}  
        \STATE Sample $N$ QA triplets $\{(q_i,c_i,o_i^{\star})\}_{i=1}^N$ from the examiner policy ${\pi_\phi^E}$

        \STATE Update ${\pi_\phi^E}(\cdot)$ with Eq.~(\ref{Eq:examiner_loss})
    \ENDFOR
    \STATE \
    \STATE \textit{\textcolor{gray}{\# Generate Training Data For Student}}
    \STATE Generate $\mathcal{D}=\{(q_i,c_i,o_i^{\star})\}_{i=1}^{N_{\mathcal{D}}}$ from ${\pi_\phi^E}(\cdot)$
    \STATE \
    \STATE \textit{\textcolor{gray}{\# Updating Student and Teacher Model}}
    \FOR{step $k \leftarrow 1$ to $W$}
        \STATE Sample a batch $\mathcal{D}_b$ from $\mathcal{D}$
        \FOR{each QA triplets $(q,c,o^{\star})$ in $\mathcal{D}_b$}
        \STATE Sample $G$ rollouts $\{y_i\}_{i=1}^G$ from the student policy $ {\pi_{\theta}^S}(\cdot | q)$
        \ENDFOR

        \STATE Update $ {\pi_{\theta}^S}$ with Eq.~(\ref{Eq:student_loss}) by teacher guidance; \textbf{Then} Soft Update ${\pi_\psi^T}$ with Eq.~(\ref{Eq:teacher_update})
    \ENDFOR
\ENDFOR
\RETURN Student model ${\pi_\theta^S}$
\end{algorithmic}
\label{algo:training}
\end{algorithm}

\subsection{The Co-Evolution Procedure of {$\boldsymbol{\pi}$-Play }}
In summary, we present $\pi$-Play, a data-free self-evolution framework that jointly optimizes the examiner, teacher, and student models (Fig.~\ref{fig:framework_compare}). In each iteration, the examiner generates QA pairs together with QCP, and is trained via student-derived difficulty feedback to produce challenging yet solvable questions. The student improves its search and reasoning abilities through training on QA pairs generated by examiner, while the teacher leverages the QCP as additional context to provide token-level guidance to the student. After each update, the teacher is softly aligned with the student, enabling both models to co-evolve. This alternating optimization loop forms a symbiotic feedback cycle in which stronger students drive the examiner toward more challenging questions, while teacher guidance accelerates student improvement. The training process is summarized in Algorithm~\ref{algo:training}.

\begin{table*}[t]
\centering
\caption{\textbf{Learning dynamics of \ours with increasing iterations.} }
\resizebox{0.95\linewidth}{!}{
\setlength{\extrarowheight}{0pt} 
\begin{tabular}{c|lccccccc|c}
\toprule
  &  & \multicolumn{3}{c}{\textbf{General QA}} & \multicolumn{4}{c}{\textbf{Multi-Hop QA}} &  \\
  \cmidrule(lr){3-5} \cmidrule(lr){6-9}
\textbf{} & {\textbf{Methods}}  & \textbf{NQ}    & \textbf{TriviaQA} & \textbf{PopQA} & \textbf{HotpotQA} & \textbf{2WikiMQA} & \textbf{MuSiQue} & \textbf{Bamboogle} & \textbf{Total} \\ 

\midrule
\rowcolor{lightorange}\multicolumn{10}{l}{\textbf{\textit{Qwen3-4B}}}   \\
\multirow{2}{*}{Iter1}
&Dr.Zero         & 36.1 & 58.7 & 42.8 & 28.6 & 25.9 & 6.2 & 21.6 & 219.9 \\
&\ours           & \textbf{38.6} & \textbf{59.4} & \textbf{43.5} & \textbf{32.2} & \textbf{31.4} & \textbf{10.7} & \textbf{32.8} & \textbf{248.6} \\

\cmidrule(l){2-10} 
\multirow{2}{*}{Iter2}
&Dr.Zero         & \textbf{40.6} & \textbf{60.5} & \textbf{44.4} & 30.8 & 29.1 & 8.0 & 28.8 & 242.2 \\
&\ours           & 38.1 & 58.4 & 42.3 & \textbf{32.3} & \textbf{32.5} & \textbf{11.2} & \textbf{35.2} & \textbf{250.0} \\

\cmidrule(l){2-10} 
\multirow{2}{*}{Iter3}
&Dr.Zero    & 40.4 & 61.2 & 44.3 & 31.0 & 29.3 & 7.8 & \textbf{31.2} & 245.2 \\
&\ours      & \textbf{41.1} & \textbf{62.0} & \textbf{44.8} & \textbf{32.5} & \textbf{31.5} & \textbf{9.0} & 28.8 & \textbf{249.7} \\
\toprule

\rowcolor{lightorange}\multicolumn{10}{l}{\textbf{\textit{Qwen3-4B-Instruct}}}   \\
\multirow{2}{*}{Iter1}
&Dr.Zero         & 39.8 & 61.7 & 42.6 & 35.4 & 27.4 & 11.7 & 36.0 & 254.6 \\
&\ours           & \textbf{40.7} & \textbf{62.1} & \textbf{42.9} & \textbf{36.7} & \textbf{30.1} & \textbf{13.2} & \textbf{39.2} & \textbf{264.9} \\

\cmidrule(l){2-10} 
\multirow{2}{*}{Iter2}
&Dr.Zero         & 39.3 & 61.7 & 43.2 & \textbf{38.2} & 28.4 & \textbf{14.2} & 38.4 & 263.4 \\
&\ours           & \textbf{41.2} & \textbf{62.5} & \textbf{43.5} & 37.8 & \textbf{32.5} & \textbf{14.2} & \textbf{40.0} & \textbf{271.7}   \\

\cmidrule(l){2-10} 
\multirow{2}{*}{Iter3}
&Dr.Zero         & 41.3 & 61.6 & \textbf{44.0} & 38.4 & 29.8 & \textbf{14.4} & 36.8 & 266.3 \\
&\ours           & \textbf{41.8} & \textbf{63.2} & \textbf{44.0} & \textbf{38.5} & \textbf{32.3} & 13.4 & \textbf{44.0} & \textbf{277.2} \\

\toprule

\rowcolor{lightorange}\multicolumn{10}{l}{\textbf{\textit{Qwen3-8B}}}   \\
\multirow{2}{*}{Iter1}
&Dr.Zero         & 36.7 & 59.5 & 42.9 & 32.1 & 26.9 & 9.5 & 18.4 & 226.0 \\
&\ours           & \textbf{38.2} & \textbf{59.7} & \textbf{43.4} & \textbf{34.1} & \textbf{33.2} & \textbf{10.9} & \textbf{35.2} & \textbf{254.7} \\

\cmidrule(l){2-10} 
\multirow{2}{*}{Iter2}
&Dr.Zero         & 36.7 & 60.0 & 43.2 & 36.6 & \textbf{35.4} & \textbf{12.7} & \textbf{38.4} & 263.0 \\
&\ours           & \textbf{42.9} & \textbf{63.2} & \textbf{46.4} & \textbf{37.7} & 34.7 & 12.4 & 36.0 & \textbf{273.3} \\

\cmidrule(l){2-10} 
\multirow{2}{*}{Iter3}
&Dr.Zero   & 42.1 & 62.5 & 45.8 & 36.6 & \textbf{34.8} & \textbf{13.1} & \textbf{40.0} & 274.9 \\
&\ours     & \textbf{43.0} & \textbf{64.6} & \textbf{47.2} & \textbf{38.9} & 34.2 & 12.4 & \textbf{40.0} & \textbf{280.3} \\

\bottomrule
\end{tabular}
}
\label{tab:iteration_compare}
\end{table*}

\section{Experiments}
\subsection{Setup}
\paragraph{Datasets \& Models.}
We conduct experiments on three models from the Qwen-3 series \citep{yang2025qwen3technicalreport}. The base models are Qwen3-4B,  Qwen3-4B-Instruct, and Qwen3-8B. More details on experimental setups and training procedures can be found in Appendix~\ref{app:Implementation}. We evaluate \ours primarily on three one-hop benchmarks NQ \citep{kwiatkowski-etal-2019-natural}, TriviaQA \citep{JoshiTriviaQA2017}, PopQA \citep{mallen-etal-2023-trust}, as well as four multi-hop QA benchmarks, including 
HotpotQA \citep{yang2018hotpotqadatasetdiverseexplainable}, 
2WikiMQA \citep{ho2020constructingmultihopqadataset}, 
MuSiQue \citep{trivedi2022musiquemultihopquestionssinglehop}, and 
Bamboogle \citep{press2023measuringnarrowingcompositionalitygap}.

\paragraph{Baselines \& Evaluation.} 
To demonstrate the efficacy of $\pi$-Play, we compare it against a variety of baseline search agents: (1) \textit{training-free}: ReAct \citep{yao2023react}; (2) \textit{supervised RL}: Search-R1 \citep{jin2025searchr1trainingllmsreason} and ToolForge \citep{chen2025toolforgedatasynthesispipeline}, and (3) \textit{ self-play}: Dr.Zero \citep{yue2026drzeroselfevolvingsearch} and SQLM* \citep{chen2025selfquestioninglanguagemodels,yue2026drzeroselfevolvingsearch}. All models are evaluated using the exact match scores with identical search engine (E5-base \citep{wang2024textembeddingsweaklysupervisedcontrastive}) and corpus settings (English Wikipedia dump \citep{karpukhin2020densepassageretrievalopendomain}). We report the performance of the checkpoint from its best-performing training iteration (step).


\begin{figure}[t]
    \centering
    \includegraphics[width=1\linewidth]{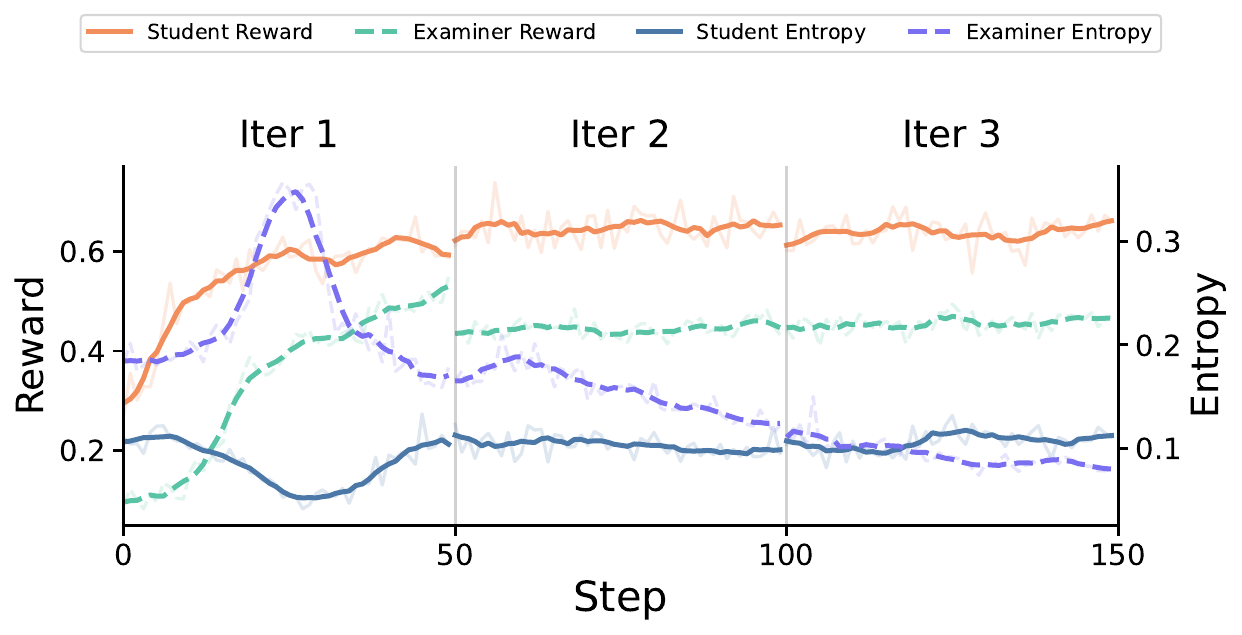}
    \caption{\textbf{Iterative reward and entropy dynamics of the examiner and student in \ours with Qwen3-4B-Instruct}. Both reward and entropy reach a converged state by Iteration 3.}
    \label{fig:reward_entropy}
\end{figure}

\subsection{Main Results}
We first analyze the main evaluation results as reported in Table~\ref{tab:main_result} and derive several key observations from them: (1) \textit{Strong Overall Performance: }\ours achieves substantial improvements in overall performance over base LLM (e.g., ReAct), demonstrating strong robustness and generalization across diverse task types and model scales. The results further demonstrate the effectiveness and superiority of the multi-agent self-evolution framework for search agents. (2) \textit{\ours surpasses supervised RL methods.} \ours delivers strong performance without using any training data. In terms of average performance, it surpasses the Search-R1 by 6.3\%, 4.2\%, and 15.4\% on Qwen3-4B, Qwen3-4B-Instruct, and Qwen3-8B, respectively. This self-evolution framework demonstrates greater performance gains when instantiated with a stronger base LLM. (3) \textit{\ours surpasses self-play methods.} 
\ours outperforms the self-play methods (SQLM* and Dr.Zero) across multiple model scales, benefiting from the additional guidance provided by the teacher to the student. Notably, the performance gains are even more substantial on multi-hop benchmarks, where complex multi-step reasoning is required. We attribute this advantage to token-level credit assignment from the teacher, which provides finer-grained and effective supervision for long-horizon reasoning.

\subsection{Training Dynamics}
To better understand the self-evolving dynamics of $\pi$-Play, we analyze model performance across training iterations, with detailed comparisons with Dr.Zero summarized in  Table~\ref{tab:iteration_compare}. These results lead to several key observations: (1) Across all three iterations, the student model shows a steady upward performance trend, consistently outperforming Dr.Zero after every iteration. This highlights the effectiveness of the examiner–teacher–student interplay and suggests that collaborative self-evolution is more effective than standard self-play by enabling more efficient information sharing among agents. (2) After the first training iteration, the student in \ours achieves substantial gains in both search and reasoning abilities. In terms of performance, it already matches or even surpasses the converged performance of Dr.Zero after three iterations, demonstrating the superior evolutionary efficiency. (3) After the second iteration, the improvement trends begin to vary across model sizes. While Qwen3-4B-Instruct and Qwen3-8B continue to show modest gains, Qwen3-4B drops slightly from 250 to 249.7, suggesting that performance has started to plateau. Beyond this point, further iterations bring only marginal or no additional improvements across model sizes. In summary, these dynamics validate the design of $\pi$-Play, demonstrating that the introduction of the privileged teacher brings significant benefits.

\section{Further Analysis}
In this section, we analyze the importance of the QCP, the co-evolution dynamic of the examiner, and the search behavior of the student. Further experiments, including extensive ablation studies, training cost are provided in Appendix~\ref{app:further_analysis}.

\begin{table*}[t]
\centering
\caption{\textbf{Ablation study on the question construction path (QCP) with Qwen3-4B-Instruct.} \ours w/o Distillation denotes the variant of \ours without teacher-guided distillation loss. Variants of the form \ours w/ [Privileged Info] replace QCP with different forms of privileged information for self-distillation.}
\resizebox{1\linewidth}{!}{
\setlength{\extrarowheight}{-1pt} 
\begin{tabular}{@{}lcccccccc@{}}
\toprule
   & \multicolumn{3}{c}{\textbf{General QA}} & \multicolumn{4}{c}{\textbf{Multi-Hop QA}} &  \\
\cmidrule(lr){2-4} \cmidrule(lr){5-8}
\multirow{2}{*}{}    & \textbf{NQ}    & \textbf{TriviaQA} & \textbf{PopQA} & \textbf{HotpotQA} & \textbf{2WikiMQA} & \textbf{MuSiQue} & \textbf{Bamboogle} & \textbf{Total} \\ 
\midrule
\ours w/o Distillation          & 41.2          & 62.3             & 43.6          & 37.7             & 29.0             & 13.3            & 38.4              & 265.5 \\
\ours w/ GT     & 42.0 & 62.0 & \textbf{44.5} & 37.7 & 29.4 & 13.3 & 35.2 & 264.1 \\
\ours w/ GT+Hop     & \textbf{42.1} & 61.9 & 42.9 & 37.7 & 29.1 & \textbf{14.0} & 39.2 & 266.9 \\
\ours w/ Partial QCP & 41.6 & 62.1 & 43.3 & 38.3 & 31.0 & 13.6 & 43.2 & 273.1 \\
\rowcolor{lightpurple}\ours w/ QCP (Ours)    & 41.8 & \textbf{63.2} & 44.0 & \textbf{38.5} & \textbf{32.3} & 13.4 & \textbf{44.0} & \textbf{277.2} \\
\bottomrule
\label{tab:PI_ablation}
\end{tabular}
}
 \vspace{-1.5em}
\end{table*}

\subsection{Ablation Study on the Question Construction Path}
As shown in Table~\ref{tab:PI_ablation}, we further evaluate the effectiveness of QCP as privileged information by comparing it with several alternative forms of teacher-side privileged context, including the ground-truth answer alone (\ours w/ GT), the ground-truth answer combined with the question hop count (\ours w/ GT+HOP), and a partial QCP obtained by randomly truncating half of the original QCP (\ours w/ Partial QCP). Among these variants, using GT yields the worst performance, which is nearly comparable to the variant without a teacher model (\ours w/o Distillation). This is because GT contains only the final answer and provides little information about the underlying logic used to construct the question, making it difficult for the teacher to provide effective guidance to the student. In contrast, the full QCP achieves the best performance, while Partial QCP performs second best. Moreover, Partial QCP still substantially outperforms both GT and GT+HOP, further highlighting the effectiveness of QCP as privileged information.

\begin{table}[h]
\centering
\caption{\textbf{Co-evolutionary dynamic of the examiner.} We report the student's accuracy on the datasets generated by the examiner at different steps.}
\resizebox{1\linewidth}{!}{
\setlength{\extrarowheight}{0pt} 
\begin{tabular}{@{}lccc}
\toprule
 & \multicolumn{3}{c}{{\textbf{Performance of Evaluated Model}}} \\
\cmidrule(l){2-4}    & Base Model    & Student(step 50) & Student(step 100) \\ 
\midrule
$\mathcal{D}_{\text{step 50}}$      & 27.5  & 57.1 & 58.4 \\
$\mathcal{D}_{\text{step 100}}$      & 27.0  & 51.9 & 57.3\\
$\mathcal{D}_{\text{step 150}}$      & 22.9  & 45.1 & 53.3   \\

\bottomrule

\label{tab:question_difficulty}
\end{tabular}
}
\end{table}
 \vspace{-2em}

\subsection{Evolution of Question Difficulty} 
To understand the co-evolutionary dynamic of the examiner, we examined how the tasks it generated changed across iterations. After each of the training iterations, we sampled 2000 questions from its policy, creating three distinct evaluation sets: $\mathcal{D}_{\text{step 50}}$, $\mathcal{D}_{\text{step 100}}$, and $\mathcal{D}_{\text{step 150}}$. As shown in Table~\ref{tab:question_difficulty}, the examiner generates progressively more challenging questions as training proceeds. This is evidenced by the performance of a fixed solver on these evolving question sets: for instance, the static student (Step 50) drops from 57.1 on $\mathcal{D}_{\text{step 50}}$ to 45.1 on $\mathcal{D}_{\text{step 150}}$. This suggests that examiner successfully increases task difficulty over the course of training.

\subsection{Search Behavior Analysis}
To assess the effect of QCP-guided self-distillation on search behavior, we quantitatively analyze search count and query redundancy across seven QA benchmarks. As shown in Table~\ref{tab:redundant}, compared with conventional self-play methods such as SQLM* and Dr.Zero, \ours achieves higher accuracy with fewer search actions, indicating more efficient behavior. The same trend holds for query redundancy. Fine-grained supervision leads to lower redundancy and more effective queries.

\begin{table}[h]
\centering
\caption{\textbf{Quantitative analysis of search count and query redundancy with Qwen3-4B-Instruct.} We report the average number of search actions and the average query redundancy across seven QA benchmarks (NQ, TriviaQA, PopQA, HotpotQA, 2WikiMQA, MuSiQue, and Bamboogle). }
\label{tab:redundant}
\resizebox{1\linewidth}{!}{
\setlength{\extrarowheight}{2pt}
\begin{tabular}{c|ccc}
\toprule
 & \textbf{Accuracy $\uparrow$} & \textbf{Search Count $\downarrow$} & \textbf{Query Redundancy $\downarrow$} \\
\midrule
\textbf{SQLM*} & 35.7 & 2.3 & 0.43 \\
\textbf{Dr.Zero} & 38.0 & 2.4 & 0.48 \\
\rowcolor{lightpurple}\textbf{\blackours} & \textbf{39.6} & \textbf{1.9} & \textbf{0.37} \\
\bottomrule
\end{tabular}
}
\end{table}

\section{Conclusion}

In this work, we propose \ours, a novel self-evolution framework that automatically generates training tasks through self-play, which provides high-quality privileged information for self-distillation at low cost and at scale. Our key insight is that the QCP generated during self-play constitutes an intrinsic form of privileged information that can be transformed into dense token-level supervision through self-distillation. Extensive experiments demonstrate that \ours outperforms fully supervised search agents and enables more efficient LLM self-evolution with no need for external labels or a stronger teacher.

\section{Limitations}

While \ours provides a new framework for multi-agent self-evolution, our current experiments instantiate it with a foundational self-distillation pipeline. This design helps measure the effectiveness of the QCP-guided self-distillation, but it may not fully exploit recent advances in self-distillation. Consequently, the evolutionary efficiency and supervision quality of \ours could be further enhanced by incorporating more advanced on-policy distillation and self-distillation techniques. Exploring such integrations is an important direction for future work.

\bibliography{reference}
\clearpage
\appendix
\section{Related Work}
\subsection{Deep Search Agents}
Deep search agents leverage the reasoning capabilities of large language models and external search engines to perform multi-turn retrieval and analysis for complex questions, emerging as a promising paradigm for information acquisition. Recent work has leveraged RL to further enhance both reasoning and access to up-to-date knowledge, enabling LLMs to tackle complex tasks more effectively \citep{shao2024deepseekmathpushinglimitsmathematical,deepseekai2025deepseekr1incentivizingreasoningcapability}. Some agentic RL works, including Search-R1 \citep{jin2025searchr1trainingllmsreason}, R1-Searcher \citep{song2025r1searcherincentivizingsearchcapability}, DeepResearcher \citep{zheng-etal-2025-deepresearcher}, and ZeroSearch \citep{sun2025zerosearchincentivizesearchcapability}, further enhance question-answering capabilities but remain constrained by limited training data. To scale agentic RL, some pipelines \citep{wu2025webdancer, li2025websailor, gao2025turnsunlockinglonghorizonagentic} employ offline question-synthesis strategies, yet they do not explicitly couple task generation with the evolving capability of the solver. In contrast, self-play enables search agents to jointly generate and solve tasks without human annotations, reducing reliance on manual supervision and allowing agentic RL to scale to broader scenarios \citep{huang2026rzeroselfevolvingreasoningllm, lu2025searchselfplaypushingfrontier, yue2026drzeroselfevolvingsearch}.
\subsection{Self-play for LLMs}
Self-play enables LLMs to autonomously improve their reasoning and problem-solving capabilities by iteratively generating tasks and learning from their own experiences \citep{liu2026spiralselfplayzerosumgames,huang2026rzeroselfevolvingreasoningllm, lu2025searchselfplaypushingfrontier,liu2025spiceselfplaycorpusenvironments, yue2026drzeroselfevolvingsearch}. Early approaches leverage the LLM as both generator and evaluator, refining its policy without human supervision \citep{openai2021asymmetricselfplayautomaticgoal,chen2024selfplayfinetuningconvertsweak}. For instance, self-rewarding LLMs employ iterative training loops where the model judges its own outputs to construct preference data for optimization \citep{pmlr-v235-yuan24d}. 
More recent frameworks, such as R-Zero \citep{huang2026rzeroselfevolvingreasoningllm}, SSP \citep{lu2025searchselfplaypushingfrontier}, and Dr.Zero \citep{yue2026drzeroselfevolvingsearch}, typically involve only two roles: an \textit{examiner}, which generates QA pairs, and a \textit{student}, which is optimized via outcome-based RL on these self-generated tasks. These methods typically rely on sparse outcome rewards and do not exploit the construction paths produced during question generation. Such coarse-grained feedback makes it difficult to distinguish effective from ineffective tool-use behaviors, leading to inefficient credit assignment and slow policy improvement. In contrast, \ours incorporates these construction paths into student optimization as token-level supervision.

\subsection{Privileged Information Self-distillation for LLMs}
Self-distillation is a training paradigm that enables a student model to improve by learning from its own generated outputs. In this process, a teacher model evaluates students' rollouts and provides supervision signals at the token level to guide students. Unlike on-policy distillation \citep{agarwal2024onpolicy,gu2024minillm,yue2025does,lu2025onpolicydistillation}, self-distillation does not rely on a larger external teacher. Instead, the teacher typically shares the same architecture and scale as the student,  but is augmented with privileged information to provide reliable supervision \citep{2026reinforcementlearningselfdistillation, shenfeld2026selfdistillationenablescontinuallearning, zhao2026selfdistilledreasoneronpolicyselfdistillation, ye2026onpolicycontextdistillationlanguage, penaloza2026privilegedinformationdistillationlanguage}. Such dense supervision has been shown to effectively enhance the student model's learning efficiency \citep{2026reinforcementlearningselfdistillation, shenfeld2026selfdistillationenablescontinuallearning,ye2026onpolicycontextdistillationlanguage, penaloza2026privilegedinformationdistillationlanguage}. However, obtaining high-quality privileged information is often nontrivial. In several prior works, privileged information is typically constructed with the help of human experts or stronger models \citep{shenfeld2026selfdistillationenablescontinuallearning,zhao2026selfdistilledreasoneronpolicyselfdistillation,ye2026onpolicycontextdistillationlanguage}, including expert demonstrations \citep{penaloza2026privilegedinformationdistillationlanguage,shenfeld2026selfdistillationenablescontinuallearning} and prior knowledge \citep{ye2026onpolicycontextdistillationlanguage,sang2026crispcompressedreasoningiterative}, which limits the scalability of self-distillation. In \ours, the privileged signal provided to the teacher model is derived from the task-construction process in self-play, thereby avoiding dependence on externally provided privileged information.

\section{Implementation}
\label{app:Implementation}
In our experiments, we implement \ours through alternating optimization over the examiner and the student-teacher modules. Following prior self-play work, the examiner generates 1-, 2-, 3-, and 4-hop questions with a default ratio of 4:3:2:1. In each iteration (i.e., iter1, iter2, and iter3), we first train the examiner for 50 steps, then use it to generate QA data from the corresponding prompts, and subsequently train the student on the synthesized data for another 50 steps. Meanwhile, the teacher is soft-updated at every student training step. Consistent with prior self-play settings, we run only three iterations in total, yielding 150 total training steps for each model, which is substantially fewer than baselines such as Search-R1. Throughout training, we adopt a decayed $\lambda$ schedule to gradually  weaken teacher guidance, allowing the student to progressively move toward regions with higher EM reward. For Qwen3-4B and Qwen3-8B, we set $\lambda$ to 0.03, 0.003, and 0.002 across the three iterations, respectively. For Qwen3-4B-Instruct, we set $\lambda$ to 0.1, 0.03, and 0.03. The format reward $r^f$ for the examiner is set in the same way as the proposer's format reward in Dr.Zero \citep{yue2026drzeroselfevolvingsearch}. Full hyperparameter details are reported in Table~\ref{tab:examiner-hyperparameter}, Table~\ref{tab:student-hyperparameter}, and Table~\ref{tab:teacher-hyperparameter}.

\begin{table}[h] 
\centering
\caption{\textbf{Examiner hyperparameter settings.}}
\vspace{5pt}

{
\begin{tabular}{lccccc}
    \toprule
    Steps & \multicolumn{5}{c}{\texttt{50}} \\
    Optimizer & \multicolumn{5}{c}{\texttt{AdamW}} \\
    Optimizer Momentum & \multicolumn{5}{c}{\texttt{$\beta_1$, $\beta_2$ = 0.9, 0.999}} \\
    Warmup Ratio & \multicolumn{5}{c}{\texttt{0.03}} \\
    Weight Decay & \multicolumn{5}{c}{\texttt{0.01}} \\
    Learning Rate & \multicolumn{5}{c}{\texttt{5e-7, 1e-6}} \\
    Max Gradient Norm & \multicolumn{5}{c}{\texttt{0.1, 1.0}} \\
    Group size  & \multicolumn{5}{c}{\texttt{1}} \\
    Reward size & \multicolumn{5}{c}{\texttt{5}} \\
    KL-Div & \multicolumn{5}{c}{\texttt{0}} \\
    Total Train Batch Size & \multicolumn{5}{c}{\texttt{256}} \\
    LR Scheduler & \multicolumn{5}{c}{\texttt{Constant with Warmup}} \\
    Precision (WA) & \multicolumn{5}{c}{\texttt{BF16-mixed}} \\
    Max Turn in Rollout & \multicolumn{5}{c}{\texttt{5}} \\
    Max Sequence Length & \multicolumn{5}{c}{\texttt{4096}} \\
    \bottomrule
\end{tabular}
}
\label{tab:examiner-hyperparameter}
\end{table}

\begin{table}[h] 
\centering
\caption{\textbf{Student hyperparameter settings.}}
\vspace{5pt}

{
\begin{tabular}{lccccc}
    \toprule 
    Steps & \multicolumn{5}{c}{\texttt{50}} \\
    Optimizer & \multicolumn{5}{c}{\texttt{AdamW}} \\
    Optimizer Momentum & \multicolumn{5}{c}{\texttt{$\beta_1$, $\beta_2$ = 0.9, 0.999}} \\
    Warmup Ratio & \multicolumn{5}{c}{\texttt{0.03}} \\
    Weight Decay & \multicolumn{5}{c}{\texttt{0.01}} \\
    Learning Rate & \multicolumn{5}{c}{\texttt{1e-6}} \\
    Max Gradient Norm & \multicolumn{5}{c}{\texttt{0.1, 1.0}} \\
    Group size in GRPO & \multicolumn{5}{c}{\texttt{5}} \\
    KL-Div in GRPO & \multicolumn{5}{c}{\texttt{0, 0.001}} \\
    $\epsilon$ in GRPO & \multicolumn{5}{c}{\texttt{0.2}} \\
    Total Train Batch Size & \multicolumn{5}{c}{\texttt{256}} \\
    LR Scheduler & \multicolumn{5}{c}{\texttt{Constant with Warmup}} \\
    Precision (WA) & \multicolumn{5}{c}{\texttt{BF16-mixed}} \\
    Max Turn in Rollout & \multicolumn{5}{c}{\texttt{5}} \\
    Max Sequence Length & \multicolumn{5}{c}{\texttt{8192}} \\
    \bottomrule
\end{tabular}
}
\label{tab:student-hyperparameter}
\end{table}

\begin{table}[h] 
\centering
\caption{\textbf{Teacher hyperparameter settings.}}
\vspace{5pt}

{
\begin{tabular}{lccccc}
    \toprule 
    Precision (WA) & \multicolumn{5}{c}{\texttt{BF16-mixed}} \\
    Soft Update Weight $\tau$ & \multicolumn{5}{c}{\texttt{0.05}} \\
    Max Sequence Length & \multicolumn{5}{c}{\texttt{8192}} \\
    \bottomrule
\end{tabular}
}
\label{tab:teacher-hyperparameter}
\end{table}

\begin{table*}[h]
\centering
\caption{\textbf{Ablation study of $\lambda$.} \textbf{Bold} value indicates the top-performing result, while \uline{underline} value denotes the second-best.}
\label{tab:ablation_lambda}
\resizebox{1\linewidth}{!}{
\setlength{\extrarowheight}{0pt} 
\begin{tabular}{@{}l|cccccccc}
\toprule
   & \multicolumn{3}{c}{\textbf{General QA}} & \multicolumn{4}{c}{\textbf{Multi-Hop QA}} &  \\
   \cmidrule(lr){2-4} \cmidrule(lr){5-8}
\multirow{2}{*}{\textbf{$\boldsymbol{\lambda}$}}    & \textbf{NQ}    & \textbf{TriviaQA} & \textbf{PopQA} & \textbf{HotpotQA} & \textbf{2WikiMQA} & \textbf{MuSiQue} & \textbf{Bamboogle} & \textbf{Total} \\ 
\cmidrule(l){2-9} & \multicolumn{8}{c}{ Iter 1} \\ \midrule
\textbf{0.100}         & 40.1          & \underline{61.8}             & 42.5          & \underline{35.5}             & 30.1             & \underline{12.8}            & \underline{40.8}              & \underline{263.6}            \\
\textbf{0.050}         & 39.8 & \textbf{62.1}             & 42.2          & 34.9             & \textbf{31.4}             & 11.8   & \textbf{41.6}     & 263.8            \\
\textbf{0.030}        & 40.1          & 61.1             & \underline{42.7}          & 33.7             & 30.4             & 9.85            & 40.0              & 257.9            \\ 
\textbf{0.010}        & 38.5          & 61.6             & 42.5          & 35.4             & 26.6             & 12.4            & 35.2              & 252.2            \\ 
\textbf{0.005}        & 40.1          & 61.1             & 42.0          & 34.9             & \underline{31.1}             & 10.6            & 32.0              & 251.8            \\ 
\textbf{0.000}        & \textbf{40.8}          & 61.3             & 42.3          & 34.3             & 27.9             & 11.1            & 31.2              & 248.9            \\ 
\rowcolor{lightpurple}\textbf{Decay}        & \underline{40.7}          & \textbf{62.1}             & \textbf{42.9}          & \textbf{36.7}             & 30.1             & \textbf{13.2}            & 39.2              & \textbf{264.9}           \\

\cmidrule(l){2-9} 
\multicolumn{1}{c}{\textbf{}} & \multicolumn{8}{c}{ Iter 2} \\ \midrule
\textbf{0.100}         & 40.1          & 61.5             & 42.5          & 36.0             & 30.5             & \underline{13.8}            & \textbf{40.0}              & 264.4            \\
\textbf{0.050}         & 40.6 & 62.2             & 43.6          & 36.2             & \underline{31.3}             & 12.8   & 35.2     & 261.9            \\
\textbf{0.030}        & 40.9          & 61.7             & 43.2          & \underline{37.4}             & 28.9             & 13.6            & \underline{38.4}              & 264.1            \\
\textbf{0.010}        & \underline{41.0}          & 62.3             & \underline{43.7}          & 37.2             & 30.4             & 12.8            & 36.8              & \underline{264.2}            \\
\textbf{0.005}        & \textbf{41.2}          & \textbf{62.6}             & \textbf{44.0}          & 37.3             & 28.1             & 13.0            & 36.8              & 263.0            \\
\textbf{0.000}        & \textbf{41.2}          & 61.6             & 43.4          & 36.3             & 26.1             & \underline{13.8}            & 36.0              & 258.4            \\
\rowcolor{lightpurple}\textbf{Decay}        & \textbf{41.2}          & \underline{62.5}             & 43.5          & \textbf{37.8}             & \textbf{32.5}             & \textbf{14.2}            & \textbf{40.0}              & \textbf{271.7}            \\

\cmidrule(l){2-9} 
\multicolumn{1}{c}{\textbf{}} & \multicolumn{8}{c}{ Iter 3} \\ \midrule
\textbf{0.100}         & 40.8          & 61.7             & 42.3          & 35.6             & 31.0             & 13.4            & \underline{40.0}              & 264.8            \\
\textbf{0.050}         & 41.0          & 61.3             & 42.8          & 35.7             & 30.5             & \underline{14.0}            & 36.0              & 261.3            \\
\textbf{0.030}        & \textbf{42.5}          & 62.0             &\underline{44.3}          & 38.0             & 29.8             & \textbf{14.2}            & 37.6              & \underline{268.4}            \\ 
\textbf{0.010}        & \textbf{42.5}          & 62.1             & \textbf{44.5}          & 36.5             & \underline{32.0}             & 12.0            & 36.0              & 265.6            \\ 
\textbf{0.005}        & \underline{42.4}          & \underline{62.4}             & 43.9          & 37.6             & 28.6             & 13.8            & 38.4              & 267.1            \\
\textbf{0.000}        & 41.2          & 62.3             & 43.6          & 37.7             & 29.0             & 13.3            & 38.4              & 265.5            \\ 
\rowcolor{lightpurple}\textbf{Decay}        & 41.8          & \textbf{63.2}             & 44.0          & \textbf{38.5}             & \textbf{32.3}             & 13.4            & \textbf{44.0}              & \textbf{277.2}            \\

\bottomrule

\end{tabular}
}
\end{table*}
\section{Further Analysis}
\label{app:further_analysis}
\subsection{Ablation Study}
For the distillation coefficient $\lambda$ in Eq.~\ref{Eq:student_loss}, we adopt a decaying schedule, as described in Appendix~\ref{app:Implementation}. Accordingly, we conduct an ablation study on Qwen3-4B-Instruct to verify whether this decaying strategy is superior to using a fixed $\lambda$ across three iterations. As shown in Table~\ref{tab:ablation_lambda}, compared with a constant $\lambda$, the decaying schedule not only maintains a higher evolution speed in the early stages, but also converges to better final performance in the later stages of training.

Moreover, we perform an ablation study on the top-$K$ parameter in Eq.~\ref{eq:distill_full}. The study is conducted on Qwen3-4B-Instruct, with $K$ selected from ${25, 50, 100, 200}$. As shown in Table~\ref{tab:ablation_k}, the results reveal that larger values of $K$ do not necessarily yield better performance. Instead, the optimal performance is achieved when $K$ is around 50.

\begin{table*}[t]
\centering
\caption{\textbf{Ablation study of $K$.} \textbf{Bold} value indicates the top-performing result}
\label{tab:ablation_k}
\resizebox{1\linewidth}{!}{
\setlength{\extrarowheight}{1pt} 
\begin{tabular}{@{}r|cccccccc}
\toprule
   & \multicolumn{3}{c}{\textbf{General QA}} & \multicolumn{4}{c}{\textbf{Multi-Hop QA}} &  \\
   \cmidrule(lr){2-4} \cmidrule(lr){5-8}
\multirow{2}{*}{\textbf{Top $K$}}    & \textbf{NQ}    & \textbf{TriviaQA} & \textbf{PopQA} & \textbf{HotpotQA} & \textbf{2WikiMQA} & \textbf{MuSiQue} & \textbf{Bamboogle} & \textbf{Total} \\ 
\cmidrule(l){2-9} & \multicolumn{8}{c}{ Iter 1} \\ \midrule
\textbf{200} & 37.8          & 61.7             & 42.1          & \textbf{37.1}             & 29.0             & 12.9            & 35.2              & 255.8            \\
\textbf{100} & 39.4          & 61.8             & 42.4         & 36.8             & 28.4             & 12.6            & 38.4              & 259.8            \\
\textbf{50}  & \textbf{40.7} & \textbf{62.1}    & \textbf{42.9} & 36.7             & \textbf{30.1}    & \textbf{13.2}    & \textbf{39.2}     & \textbf{264.9}            \\
\textbf{25}  & 40.0          & 61.2             & 41.0          & 35.1             & 29.8             & 11.5            & 36.8              & 255.4            \\

\cmidrule(l){2-9} 
\multicolumn{1}{c}{\textbf{}} & \multicolumn{8}{c}{ Iter 2} \\ \midrule
\textbf{200} & 41.4          & 61.7             & 43.2          & 35.5             & 29.8             & 11.4            & 36.0              & 259.0            \\
\textbf{100} & \textbf{41.9} & 62.2             & \textbf{44.4} & 36.8             & 28.5             & 14.0            & 36.8              & 264.6            \\
\textbf{50}  & 41.2          & \textbf{62.5}    & 43.5          & \textbf{37.8}    & \textbf{32.5}    & \textbf{14.2}    & 40.0              & \textbf{271.7}            \\
\textbf{25}  & 40.5          & 61.9             & 43.0          & 36.9             & 29.9             & 12.9            & \textbf{43.2}     & 268.3            \\

\cmidrule(l){2-9} 
\multicolumn{1}{c}{\textbf{}} & \multicolumn{8}{c}{ Iter 3} \\ \midrule
\textbf{200} & 42.3          & 62.1             & 44.6          & 37.3             & 30.3             & 13.2            & 39.2              & 269.0            \\
\textbf{100} & 41.9          & 62.4             &\textbf{44.8}          & 37.6             & 31.2             & \textbf{13.8}    & 36.8              & 268.5            \\
\textbf{50}  & 41.8          & \textbf{63.2}    & 44.0          & \textbf{38.5}    & \textbf{32.3}    & 13.4            & \textbf{44.0}     & \textbf{277.2}            \\
\textbf{25}  & \textbf{43.1} & 62.7             & 42.9 & 36.7             & 32.1             & 12.8            & 36.8              & 267.1            \\

\bottomrule

\end{tabular}
}
\end{table*}

\subsection{Training Cost}
Table~\ref{tab:time} reports the average per-step training time of different methods. 
Across various search agents, we observe that although our approach introduces an additional teacher model compared with Dr.Zero, it does not significantly increase the per-step training time.
While the teacher model performs an extra forward pass along the student’s rollout to compute token probabilities, \ours introduces almost no additional training time compared with Dr.Zero. We attribute this to the fact that Dr.Zero lacks fine-grained feedback to penalize ineffective tool-use behaviors, causing the student model to generate redundant search queries (Fig.~\ref{fig:comparecase}), thereby increasing training time. This effect occurs both during the computation of difficulty rewards in the examiner and during the student model's rollout.

The reported training times were measured on a node equipped with 8 NVIDIA H20 GPUs, and we follow the same training configurations described in Appendix~\ref{app:Implementation}.
\begin{table*}[h]
\centering
\caption{\textbf{The training time.} We analyze the per-step training time (in seconds) for each iteration under Qwen3-4B-Instruct. Although the teacher model requires an additional forward pass along the student’s rollout to compute token probabilities, \ours introduces only a small per-step training overhead of 4.3\% relative to Dr.Zero.}
\label{tab:time}
\resizebox{0.96\linewidth}{!}{
\setlength{\extrarowheight}{2pt} 
\begin{tabular}{@{}c|cc|cc|cc|cc@{}}
\toprule
 \multirow{2}{*}{}   & \multicolumn{2}{c|}{\textbf{Iter1}}    & \multicolumn{2}{c|}{\textbf{Iter2}} & \multicolumn{2}{c|}{\textbf{Iter3}} & \multicolumn{2}{c}{\textbf{Average}} \\ 
   &   Dr.Zero      &      \ours    &     Dr.Zero    &    \ours    &      Dr.Zero    &    \ours      &     Dr.Zero   &  \ours     \\
\midrule
\textbf{Examiner} &   311.6      &      311.6    &     407.8    &    526.3   &      623.4    &    529.9     &     447.6   & 455.9    \\
\textbf{Student (Rollout)}&   88.9      &      81.1    &     146.0    &    159.9   &      176.8    &    149.8     &     137.2  & 130.3    \\
\textbf{Teacher (Forward)}&   -      &      21.8    &     -    &    31.8   &      -    &    30.6     &    -  & 28.1    \\
\textbf{Student (Updating)} &   94.0      &    94.1    &     116.2    &    123.0   &      123.2    &    117.8     &     111.1   & 111.6    \\
\textbf{Total Time}  &   494.5      &      508.6    &     670.0    &    841.0   &      923.4    &    828.1     &     695.9   & 725.9    \\
\bottomrule

\end{tabular}
}
\end{table*}

\subsection{Case Study}
\begin{figure*}[t]
    \centering
    \includegraphics[width=1\linewidth]{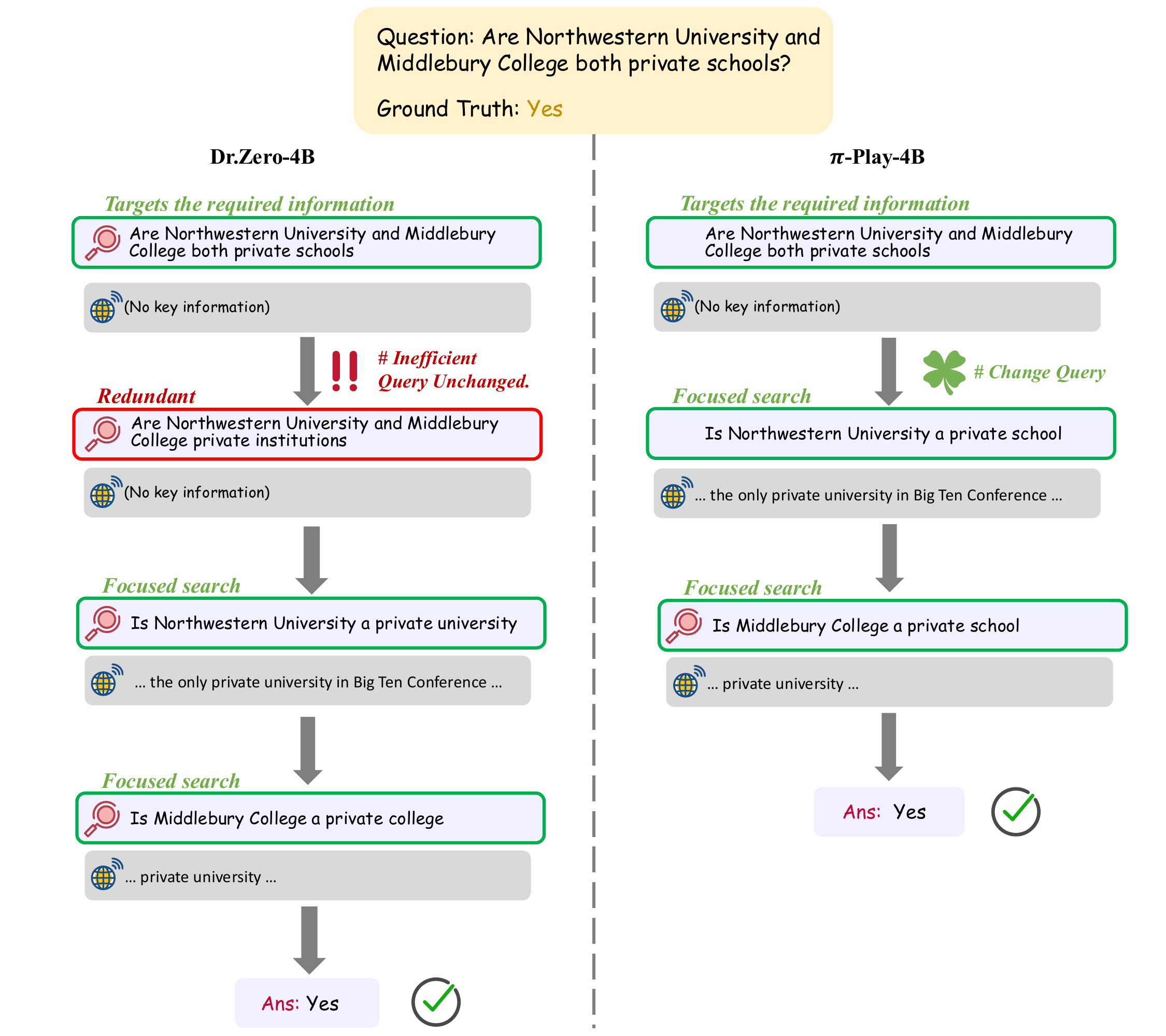}
    \caption{Side-by-side trajectories of Dr.Zero (left) and \ours (right) on the same question. Each trajectory shows multi-turn interactions with the search engine (actions, responses, and final answer). Although both Dr.Zero and \ours answer the query correctly, our method (right) uses fewer queries and reaches a logically structured answer with minimal redundancy.}
    \label{fig:comparecase}
\end{figure*}

We present QCP examples generated by the examiner during the \ours training process. The QCPs for different hop settings are presented in Fig.~\ref{fig:hop1_qcp}, Fig.~\ref{fig:hop2_qcp}, and Fig.~\ref{fig:hop4_qcp}. All cases are obtained from the model trained on Qwen3-4B-Instruct. In particular, we further compare the responses of the student model generated by the baseline Dr.Zero and our \ours method on the same question, as shown in Fig.~\ref{fig:comparecase}. Each response contains multi-turn interactions between the model and the external search engine.

\clearpage
\onecolumn
{
\begin{center} 
\begin{tcolorbox}[
  enhanced,
  breakable,                 
  colback=white,
  colframe=black!50!white,
  title={Random Document},
  width=1.0\linewidth,       
  top=-5pt,        
  bottom=-5pt,     
  fonttitle=\small
]
\lstset{style=boxedcode}
\begin{lstlisting}[
  basicstyle=\ttfamily\footnotesize,
  breaklines=true,
  breakindent=0pt,
  breakautoindent=false,
  columns=fullflexible,
  escapeinside={(*@}{@*)}
]
Now, generate a question and its answer with n = 1 hops starting from the following source document: (Title: "(*@\underline{FC Barcelona Rugby League}@*)")\nwidely speculated, but purely a hypothesis, that this branch of the successful FC Barcelona club will become members of the Rugby Football League and possibly Super League should the venture prove a success in Catalonia, as they were now playing more competitive rugby, on a more regular basis. (*@\underline{Their first ever game was against CR Sant Cugat on 13 May 2010}@*), a game which ended in a disappointing 30-20 loss. They earned their first win of the competition with a 44-30 win at home to GEiEG to give them a hope of qualifying for the finals. And, despite GEiEG winning ...
\end{lstlisting}
\end{tcolorbox}

\begin{tcolorbox}[
  enhanced,
  breakable,                 
  colback=white,
  colframe=orange,
  title={Examiner: Generate Task},
  width=1.0\linewidth,       
  top=-5pt,        
  bottom=-5pt,     
  fonttitle=\small
]
\lstset{style=boxedcode}
\begin{lstlisting}[
  basicstyle=\ttfamily\footnotesize,
  breaklines=true,
  breakindent=0pt,
  breakautoindent=false,
  columns=fullflexible,
]
<question>Which club played its first ever rugby game against CR Sant Cugat on 13 May 2010?</question>
<answer>FC Barcelona Rugby League</answer>
\end{lstlisting}
\end{tcolorbox}

\captionsetup{justification=centering} 
\captionof{figure}{\textbf{QCP example with hop = 1 provided by the examiner.}}
\label{fig:hop1_qcp}
\end{center}
\vspace{5pt}
}

\begin{center} 
\begin{tcolorbox}[
  enhanced,
  breakable,                 
  colback=white,
  colframe=black!50!white,
  title={Random Document},
  width=1.0\linewidth,       
  top=-5pt,        
  bottom=-5pt,     
  fonttitle=\small
]
\lstset{style=boxedcode}
\begin{lstlisting}[
  basicstyle=\ttfamily\footnotesize,
  breaklines=true,
  breakindent=0pt,
  breakautoindent=false,
  columns=fullflexible,
]
Now, generate a question and its answer with n = 2 hops starting from the following source document: (Title: "Luna 19") Luna 19 (a.k.a. Lunik 19) (E-8-LS series), was an unmanned space mission of the Luna program. "Luna 19" extended the systematic study of lunar gravitational fields and location of mascons (mass concentrations). It also studied the lunar radiation environment, the gamma-active lunar surface, and the solar wind. Photographic coverage via a television system was also obtained. "Luna 19" was the first of the "advanced" lunar orbiters whose design was based upon the same Ye-8-class bus used for the lunar rovers and the sample collectors. For these orbiters, designated Ye-8LS, the basic "lander stage" was topped off by a ...
\end{lstlisting}
\end{tcolorbox}

\begin{tcolorbox}[
  enhanced,
  breakable,                 
  colback=white,
  colframe=lightorange,
  coltitle=black,
  title={Examiner: Query 1},
  width=1.0\linewidth,       
  top=-5pt,        
  bottom=-5pt,     
  fonttitle=\small
]
\lstset{style=boxedcode}
\begin{lstlisting}[
  basicstyle=\ttfamily\footnotesize,
  breaklines=true,
  breakindent=0pt,
  breakautoindent=false,
  columns=fullflexible,
]
<tool_call>{"name": "search", "arguments": {"query_list": ["Luna 19 mission spacecraft class design"]}}</tool_call>
\end{lstlisting}
\end{tcolorbox}
\vspace{-5.3pt}

\begin{tcolorbox}[
  enhanced,
  breakable,                 
  colback=white,
  colframe=black!50!white,
  title={Tool Response},
  width=1.0\linewidth,       
  top=-5pt,        
  bottom=-5pt,     
  fonttitle=\small
]
\lstset{style=boxedcode}
\begin{lstlisting}[
  basicstyle=\ttfamily\footnotesize,
  breaklines=true,
  breakindent=0pt,
  breakautoindent=false,
  columns=fullflexible,
  escapeinside={(*@}{@*)}
]
(Title: "Luna 19") ... Luna 19 was the first of the "advanced" lunar orbiters whose (*@\underline{design was based upon the same Ye-8-class bus used for the lunar rovers and the sample}@*) (*@\underline{collectors}@*) ...
\end{lstlisting}
\end{tcolorbox}

\begin{tcolorbox}[
  enhanced,
  breakable,                 
  colback=white,
  colframe=lightorange,
  coltitle=black,
  title={Examiner: Query 2},
  width=1.0\linewidth,       
  top=-5pt,        
  bottom=-5pt,     
  fonttitle=\small
]
\lstset{style=boxedcode}
\begin{lstlisting}[
  basicstyle=\ttfamily\footnotesize,
  breaklines=true,
  breakindent=0pt,
  breakautoindent=false,
  columns=fullflexible,
]
<tool_call>{"name": "search", "arguments": {"query_list": ["Ye-8LS spacecraft used for lunar rovers and sample collectors"]}}</tool_call>
\end{lstlisting}
\end{tcolorbox}
\vspace{-5.3pt}

\begin{tcolorbox}[
  enhanced,
  breakable,                 
  colback=white,
  colframe=black!50!white,
  title={Tool Response},
  width=1.0\linewidth,       
  top=-5pt,        
  bottom=-5pt,     
  fonttitle=\small
]
\lstset{style=boxedcode}
\begin{lstlisting}[
  basicstyle=\ttfamily\footnotesize,
  breaklines=true,
  breakindent=0pt,
  breakautoindent=false,
  columns=fullflexible,
  escapeinside={(*@}{@*)}
]
(Title: "Luna 21") ... (*@\underline{Luna 21 (Ye-8 series)}@*) was an unmanned space mission, and its spacecraft, of the Luna program, also called Lunik 21, in 1973. The spacecraft landed on the Moon and deployed the second (*@\underline{Soviet lunar rover, "Lunokhod 2"}@*). ...
\end{lstlisting}
\end{tcolorbox}

\begin{tcolorbox}[
  enhanced,
  breakable,                 
  colback=white,
  colframe=orange,
  title={Examiner: Generate Task},
  width=1.0\linewidth,       
  top=-5pt,        
  bottom=-5pt,     
  fonttitle=\small
]
\lstset{style=boxedcode}
\begin{lstlisting}[
  basicstyle=\ttfamily\footnotesize,
  breaklines=true,
  breakindent=0pt,
  breakautoindent=false,
  columns=fullflexible,
]
<question>Which country operated the lunar rover Lunokhod 2, which was deployed by the Ye-8 series spacecraft used in the Luna 19 mission</question>
<answer>Soviet Union</answer>
\end{lstlisting}
\end{tcolorbox}

\captionsetup{justification=centering} 
\captionof{figure}{\textbf{QCP example with hop = 2 provided by the examiner}}
\label{fig:hop2_qcp}
\end{center}
\vspace{5pt}

\begin{center} 
\begin{tcolorbox}[
  enhanced,
  breakable,                 
  colback=white,
  colframe=black!50!white,
  title={Random Document},
  width=1.0\linewidth,       
  top=-5pt,        
  bottom=-5pt,     
  fonttitle=\small,
]
\lstset{style=boxedcode}
\begin{lstlisting}[
  basicstyle=\ttfamily\footnotesize,
  breaklines=true,
  breakindent=0pt,
  breakautoindent=false,
  columns=fullflexible,
]
Now, generate a question and its answer with n = 3 hops starting from the following source document: (Title: "Chris Charsley")\nseason, including the test match against Darwen which won them promotion to the First Division. He also played for Aston Villa as a guest in 1886. Charsley had a brief spell with West Bromwich Albion, whom he joined in August 1891. He made his only appearance for Albion in a 2-1 defeat to Preston North End in November of the same year, before re-joining Small Heath the following month. On 25 February 1893, he became the first Birmingham City player-and the only player who had played for the club in its Small Heath Alliance days-to represent ...
\end{lstlisting}
\end{tcolorbox}

\begin{tcolorbox}[
  enhanced,
  breakable,                 
  colback=white,
  colframe=lightorange,
  coltitle=black,
  title={Examiner: Query 1},
  width=1.0\linewidth,       
  top=-5pt,        
  bottom=-5pt,     
  fonttitle=\small
]
\lstset{style=boxedcode}
\begin{lstlisting}[
  basicstyle=\ttfamily\footnotesize,
  breaklines=true,
  breakindent=0pt,
  breakautoindent=false,
  columns=fullflexible,
]
<tool_call>{"name": "search", "arguments": {"query_list": ["Chris Charsley Birmingham City first player Small Heath Alliance"]}}</tool_call>
\end{lstlisting}
\end{tcolorbox}
\vspace{-5.3pt}

\begin{tcolorbox}[
  enhanced,
  breakable,                 
  colback=white,
  colframe=black!50!white,
  title={Tool Response},
  width=1.0\linewidth,       
  top=-5pt,        
  bottom=-5pt,     
  fonttitle=\small
]
\lstset{style=boxedcode}
\begin{lstlisting}[
  basicstyle=\ttfamily\footnotesize,
  breaklines=true,
  breakindent=0pt,
  breakautoindent=false,
  columns=fullflexible,
  escapeinside={(*@}{@*)}
]
(Title: "History of Birmingham City F.C. (1875-1965)") ... Goalkeeper (*@\underline{Chris Charsley}@*), a policeman who played as an amateur and went on to serve as chief constable of Coventry, became the (*@\underline{first Small Heath player to be capped by England}@*). ...
\end{lstlisting}
\end{tcolorbox}

\begin{tcolorbox}[
  enhanced,
  breakable,                 
  colback=white,
  colframe=lightorange,
  coltitle=black,
  title={Examiner: Query 2},
  width=1.0\linewidth,       
  top=-5pt,        
  bottom=-5pt,     
  fonttitle=\small
]
\lstset{style=boxedcode}
\begin{lstlisting}[
  basicstyle=\ttfamily\footnotesize,
  breaklines=true,
  breakindent=0pt,
  breakautoindent=false,
  columns=fullflexible,
]
<tool_call>{"name": "search", "arguments": {"query_list": ["Chris Charsley England international debut date"]}}</tool_call>
\end{lstlisting}
\end{tcolorbox}
\vspace{-5.3pt}

\begin{tcolorbox}[
  enhanced,
  breakable,                 
  colback=white,
  colframe=black!50!white,
  title={Tool Response},
  width=1.0\linewidth,       
  top=-5pt,        
  bottom=-5pt,     
  fonttitle=\small
]
\lstset{style=boxedcode}
\begin{lstlisting}[
  basicstyle=\ttfamily\footnotesize,
  breaklines=true,
  breakindent=0pt,
  breakautoindent=false,
  columns=fullflexible,
  escapeinside={(*@}{@*)}
]
(Title: "1892-93 Small Heath F.C. season") ... Chris Charsley became the first Small Heath player to play for England, and only the second, after Caesar Jenkyns for Wales, to represent his country at full international level, when he took the field (*@\underline{against Ireland on 27 February}@*) at Aston Villa\'s ground in Perry Barr, Birmingham. ...
\end{lstlisting}
\end{tcolorbox}

\begin{tcolorbox}[
  enhanced,
  breakable,                 
  colback=white,
  colframe=lightorange,
  coltitle=black,
  title={Examiner: Query 3},
  width=1.0\linewidth,       
  top=-5pt,        
  bottom=-5pt,     
  fonttitle=\small
]
\lstset{style=boxedcode}
\begin{lstlisting}[
  basicstyle=\ttfamily\footnotesize,
  breaklines=true,
  breakindent=0pt,
  breakautoindent=false,
  columns=fullflexible,
]
<tool_call>{"name": "search", "arguments": {"query_list": ["Chris Charsley England international match opponent 27 February 1893"]}}</tool_call>
\end{lstlisting}
\end{tcolorbox}
\vspace{-5.3pt}

\begin{tcolorbox}[
  enhanced,
  breakable,                 
  colback=white,
  colframe=black!50!white,
  title={Tool Response},
  width=1.0\linewidth,       
  top=-5pt,        
  bottom=-5pt,     
  fonttitle=\small
]
\lstset{style=boxedcode}
\begin{lstlisting}[
  basicstyle=\ttfamily\footnotesize,
  breaklines=true,
  breakindent=0pt,
  breakautoindent=false,
  columns=fullflexible,
  escapeinside={(*@}{@*)}
]
(Title: "1892-93 in English football") ... (*@\underline{Chris Charsley}@*) of Small Heath, who later went on to be Chief Constable of Coventry, (*@\underline{made his solitary England appearance in}@*) (*@\underline{goal}@*). ...
\end{lstlisting}
\end{tcolorbox}

\begin{tcolorbox}[
  enhanced,
  breakable,                 
  colback=white,
  colframe=orange,
  title={Examiner: Generate Task},
  width=1.0\linewidth,       
  top=-5pt,        
  bottom=-5pt,     
  fonttitle=\small
]
\lstset{style=boxedcode}
\begin{lstlisting}[
  basicstyle=\ttfamily\footnotesize,
  breaklines=true,
  breakindent=0pt,
  breakautoindent=false,
  columns=fullflexible,
]
<question>Which team did Chris Charsley play against in his only England international match?</question>
<answer>Ireland</answer>
\end{lstlisting}
\end{tcolorbox}

\captionsetup{justification=centering} 
\captionof{figure}{\textbf{QCP example with hop = 3 provided by the examiner.}}
\label{fig:hop4_qcp}
\end{center}
\vspace{5pt}

\section{Distillation Loss}
\label{app:distillation_loss}
To save GPU memory, we adopt top-\textit{K} distillation following \citep{2026reinforcementlearningselfdistillation}, where the top-\textit{K} set is defined with respect to the student distribution:
{\small
\begin{equation}
\begin{aligned}
&\mathbb{D}_{\mathrm{Distill}}\bigl({\pi_{\theta}^S}
 \,\|\, {\pi_{\psi}^{T}}) =  \frac{1}{|y_i|}\sum_{t=1}^{|y_i|} \mathrm{KL}({\pi_{\theta}^S}(\cdot \mid q, y_{i,<t}) \| \mathrm{stopgrad}({\pi_{\psi}^{T}}
(\cdot \mid q, \textcolor{red}{c}, y_{i,<t}))) \\
&= \frac{1}{|y_i|} \sum_{t=1}^{|y_i|} \sum_{\hat{y}_{i,t}\in{\pi_{\theta}^S}} \pi_{\theta}^S(\hat{y}_{i,t} \mid q, y_{i,<t}) \cdot \log \frac{\pi_{\theta}^S(\hat{y}_{i,t} \mid q, y_{i,<t})}{\mathrm{stopgrad}(\pi_{\psi}^T(\hat{y}_{i,t} \mid q, c, y_{i,<t}))}\\
&=\left (\begin{aligned}&\frac{1}{|y_i|}
        \sum_{t=1}^{|y_i|} \sum_{\hat{y}_{i,t} \in \mathrm{top}_K({\pi_{\theta}^S}
)} {\pi_{\theta}^S}
(\hat{y}_{i,t} \mid q, y_{i,<t}) \cdot \log \frac{{\pi_{\theta}^S}
(\hat{y}_{i,t} \mid q, y_{i,<t})}{\mathrm{stopgrad}({\pi_{\psi}^{T}}
(\hat{y}_{i,t} \mid q, c, y_{i,<t}))} \\
        &+ \underbrace{\frac{1}{|y_i|}
        \sum_{t=1}^{|y_i|} \sum_{\hat{y}_{i,t} \in \pi_{\theta}^S \setminus \mathrm{top}_K(\pi_{\theta}^S)} {\pi_{\theta}^S}
(\hat{y}_{i,t} \mid q, y_{i,<t}) \cdot \log \frac{{\pi_{\theta}^S}
(\hat{y}_{i,t} \mid q, y_{i,<t})}{\mathrm{stopgrad}({\pi_{\psi}^{T}}
(\hat{y}_{i,t} \mid q, c, y_{i,<t}))}}_{\text{tail}}
    \end{aligned}\right)\\
&\approx \left(\begin{aligned}&\frac{1}{|y_i|}
        \sum_{t=1}^{|y_i|} \sum_{\hat{y}_{i,t} \in \mathrm{top}_K({\pi_{\theta}^S}
)} {\pi_{\theta}^S}
(\hat{y}_{i,t} \mid q, y_{i,<t}) \cdot \log \frac{{\pi_{\theta}^S}
(\hat{y}_{i,t} \mid q, y_{i,<t})}{\mathrm{stopgrad}({\pi_{\psi}^{T}}
(\hat{y}_{i,t} \mid q, c, y_{i,<t}))} \\
        &+ \underbrace{\Big(1 - \textstyle\sum_{\hat{y}_t \in \mathrm{top}_K({\pi_{\theta}^S}
)} {\pi_{\theta}^S}
(\hat{y}_{i,t} \mid q, y_{i,<t})\Big) \cdot \log \frac{1 - \textstyle\sum_{\hat{y}_t \in \mathrm{top}_K({\pi_{\theta}^S}
)} {\pi_{\theta}^S}
(\hat{y}_{i,t} \mid q, y_{i,<t})}{\mathrm{stopgrad}\Big(1 - \textstyle\sum_{\hat{y}_t \in \mathrm{top}_K({\pi_{\theta}^S}
)} {\pi_{\psi}^{T}}
(\hat{y}_{i,t} \mid q, c, y_{i,<t})\Big)}}_{\text{tail}}
\end{aligned}\right)
\end{aligned}
\label{eq:distill_full}
\end{equation}
}

Rather than computing the full KL divergence over the entire vocabulary, we split the distillation loss into two components: the exact contribution from the student's top-\textit{K} tokens and a tail term corresponding to all remaining tokens. The tail is further approximated by collapsing the non-top-\textit{K} probability mass into a single residual term. This strategy avoids storing two full copies of vocabulary logits: one for the student and one for the teacher, thereby greatly reducing memory usage. Empirically, this approximation has negligible impact on performance, since most tokens of the vocabulary are not informative at a given time \citep{2026reinforcementlearningselfdistillation}. Further analysis of the choice of $K$ can be found in Appendix~\ref{app:further_analysis}.

\section{Prompts}
We provide the system prompts for all models in Section~\ref{app:system_prompts}, and the user prompts for the examiner, teacher, and student models in Sections~\ref{app:examiner_prompts}, \ref{app:teacher_prompts}, and \ref{app:student_prompts}, respectively.
\subsection{System Prompts}
\label{app:system_prompts}
\begin{center} 
\begin{tcolorbox}[
  enhanced,
  breakable,                 
  colback=white,
  colframe=black!50!white,
  title={System Prompt for the Examiner, Teacher and Student},
  width=1.0\linewidth,        
  top=-5pt,        
  bottom=-5pt,     
  fonttitle=\small
]
\lstset{style=boxedcode}
\begin{lstlisting}[
  basicstyle=\ttfamily\scriptsize,
  breaklines=true,
  breakindent=0pt,
  breakautoindent=false,
  columns=fullflexible,
]
You are a helpful assistant.

# Tools

You may call one or more functions to assist with the user query.

You are provided with function signatures within <tools></tools> XML tags:
<tools>
{"type": "function", "function": {"name": "search", "description": "Searches the web for relevant information based on the given query.", "parameters": {"type": "object", "properties": {"query_list": {"type": "array", "description": "A list of fully-formed semantic queries. The tool will return search results for each query.", "enum": null}}, "required": ["query_list"]}, "strict": false}}
</tools>

For each function call, return a json object with function name and arguments within <tool_call></tool_call> XML tags:
<tool_call>
{"name": <function-name>, "arguments": <args-json-object>}
</tool_call>
\end{lstlisting}
\end{tcolorbox}
\captionsetup{} 
\captionof{figure}{\textbf{System prompt for the examiner, teacher and student in $\pi$-play.} They use the same system prompt.}
\label{fig:system_prompt}
\end{center}

\subsection{Examiner Prompts}
\label{app:examiner_prompts}

\begin{center} 
\begin{tcolorbox}[
  enhanced,
  breakable,                 
  colback=white,
  colframe=black!50!white,
  title={User Prompt for the Examiner},
  width=1.0\linewidth,        
  top=-5pt,        
  bottom=-5pt,     
  fonttitle=\small
]
\lstset{style=boxedcode}
\begin{lstlisting}[
  basicstyle=\ttfamily\scriptsize,
  breaklines=true,
  breakindent=0pt,
  breakautoindent=false,
  columns=fullflexible,
]
You are an expert in question generation. Craft one challenging, deterministic question and its single, unambiguous answer based on the provided source document. The logical path must start from the document and require exactly n hops (i.e., n-1 searches) to reach the final answer.

### Definitions
1. Hop: A node in the reasoning chain. Hop 1 is the starting entity found in the document. Hop n is the final answer.

### Inputs
1. n: the exact number of hops in the reasoning chain (requiring n-1 searches).
2. Source document: the full source text.

### Process & Tools
1. Analyze the Document and Select the Starting Point
  - Read and analyze the source document.
  - Select a specific entity, event or detail explicitly mentioned in the text. This entity becomes Hop 1 (the initial clue).
2. Design the Chain Forwards
  - From Hop 1 to Hop 2: Identify a factual attribute or relation of Hop 1 that is NOT in the text but can be found via search. The result is Hop 2.
  - Iterate: Continue connecting the current Hop i to the next Hop i+1 using deterministic, verifiable relation found via search.
  - Stop at Hop n: Continue this process until you have exactly n hops. Hop n must be a single, canonical final answer.
3. Reasoning & Search Protocol
  - Always reason inside `<think> ... </think>` when you plan connections or receive new information.
  - For each hop transition that requires external information, issue search query using `<tool_call> ... </tool_call>`.
  - Search results will be provided between `<tool_response> ... </tool_response>` by the system.
4. Output Format
  - Emit a numbered sequence of EXACTLY n-1 search steps. For each search i (1 to n-1), produce:
    `<think> Reasoning step i: Identify Hop i in document/search results, formulate query to reach Hop i+1 </think>`
    `<tool_call> Query to search Hop i+1 </tool_call>`
    `[Wait for search results in <tool_response> from system]`
  - After completing all searches and arriving at Hop n, output the question and final answer:
    `<think> Final reasoning step: Confirm the chain is complete with Hop n and formulate the question </think>`
    `<question> A challenging question that provides Hop 1 (the initial clue) and asks for the final answer (Hop n) </question>`
    `<answer> The single, concise final answer (Hop n) </answer>`

### Examples
1. Example template for Hop n = 1, i.e. no search:
  `<think> [Explain how Hop 1 is selected from the source document and how the question is formulated] </think>`
  `<question> [Question based solely on the text entity Hop 1] </question>`
  `<answer> [Answer (Hop 1)] </answer>`
2. Example template for Hop n = 3, i.e. 2 searches:
  `<think> [Reasoning step 1: Find Hop 1 in the source document, formulate the query to reach Hop 2] </think>`
  `<tool_call> [Search query to find Hop 2 based on Hop 1] </tool_call>`
  `[Wait for search results in <tool_response> from system]`
  `<think> [Reasoning step 2: Reason on search results to identify Hop 2 and write the next query to find Hop 3] </think>`
  `<tool_call> [Search query to find Hop 3 based on Hop 2] </tool_call>`
  `[Wait for search results in <tool_response> from system]`
  `<think> [Final reasoning step: Confirm Hop 3 in search results and formulate the question starting from Hop 1] </think>`
  `<question> [Question starting with Hop 1, requiring the solver to find Hop 2 to eventually reach the Answer (Hop 3)] </question>`
  `<answer> [Answer (Hop 3)] </answer>`

### Critical Rules
1. Start in Document: Hop 1 must be explicitly present in the source text. Every subsequent hop must be supported by the corresponding search results.
2. Search is mandatory for n > 1: Each link between hops beyond Hop 1 must use the search engine.
3. Exact search count: Emit exactly (n-1) `<tool_call>` entries, no more, no fewer.
4. No spoilers: The question must mention only Hop 1; do not include or hint at intermediate hops.
5. Clarity: The question is self-contained; the answer is concise and direct (no extra commentary, formatting or explanation).
6. Chain integrity: Each hop must depend strictly on the previous hop. No hop should be skippable or derivable without its immediate predecessor.

Now, generate a question and its answer with n = {hops} hops starting from the following source document: {document}
\end{lstlisting}
\end{tcolorbox}
\captionsetup{} 
\captionof{figure}{\textbf{Initial instructions for the examiner in $\pi$-play.} Our prompt for examiner is developed based on Dr.Zero \citep{yue2026drzeroselfevolvingsearch}}
\label{fig:user_prompt_examiner}
\end{center}

\subsection{Teacher Prompts}
\label{app:teacher_prompts}





\begin{center} 
\begin{tcolorbox}[
  enhanced,
  breakable,                 
  colback=white,
  colframe=black!50!white,
  title={User Prompt for the Teacher (Qwen3-4B-Instruct)},
  width=1.0\linewidth,        
  top=-5pt,        
  bottom=-5pt,     
  fonttitle=\small
]
\lstset{style=boxedcode}
\begin{lstlisting}[
  basicstyle=\ttfamily\scriptsize,
  breaklines=true,
  breakindent=0pt,
  breakautoindent=false,
  columns=fullflexible,
]
You are a helpful assistant. You will be given privileged information about the reverse solution process of the question (i.e., construction process of the question). Please pretend not to know the source document used to construct the question and the ground-truth in the reverse solution process, and answer the question again with reference to the construction process of the question.

###  General Rules
1. Precise Searching: When generating search queries, ensure they are specific, semantically complete, and directly target the key information you are missing.
2. Context Retention: Remember prior conversations and search results, maintaining logical consistency across multiple rounds of searching.
3. Termination Judgment: When information are sufficient to determine the answer, immediately stop searching and output the answer.
4. Reference to Privileged Information: When outputting search actions and answers, refer to the question construction process in privileged information, but do not directly output the ground-truth in the first turn or directly use extra information from the source document.
5. Search Boundary Constraint: When searching, do not introduce information that is not contained in the question or search results, even if it exists in the source document.

###  Example (Only showcase the logical style, please do not directly imitate the specific content)
#### User
This is the reverse solution process  of the question (i.e., the process of generating the question) given by teacher:

===========================================================================
Now, generate a question and its answer with n = 3 hops starting from the following source document: (Title: "Steven Febey")
selected in the national draft. He fell just short of Robert Flower's club record of 272 games and as of 2007 only four people have played more games for the Demons After the 2002 season, Febey went to Bali with his teammates and was caught up in the bombings which devastated the island. Standing beside former Demons player Steven Armstrong and David Robbins, Febey was about to enter the Sari Club when the car bomb exploded...[omitted due to page size]

Reasoning step 1: Identify Hop 1 in the document, formulate the query to reach Hop 2
<tool_call>\n{"name": "search", "arguments": {"query_list": ["other players who played in both 1988 and 2000 AFL grand finals for Melbourne Demons"]}}\n</tool_call><|im_end|>
.....

Hop 2 in the document and search results is not explicitly mentioned, so we need to further refine the search. The document mentions David Neitz, the captain of Melbourne, and players like Shane Woewodin and Greg Healy. However, the search results point us to the 1988 and 2000 AFL Grand Finals, indicating that we need to look for a player specifically from these finals who also played for Melbourne. The final hop, or Hop 3, must be a well-known player from the Melbourne Demons who played in both 1988 and 2000 grand finals, which can be Robert Harvey based on historical knowledge.
Reasoning step 2: Identify Hop 2 in the search results and reformulate to find Hop 3
<tool_call>\n{"name": "search", "arguments": {"query_list": ["Robert Harvey Melbourne Demons 1988 2000 grand finals"]}}\n</tool_call><|im_end|>
.....

The search results confirm that Robert Harvey played for the Melbourne Demons in both 1988 and 2000, making him the well-known player who fits the criteria. Therefore, Hop 3 is "Robert Harvey". The question is formulated to start with the explicit entity in the document, "Steven Febey", and lead to the single, unambiguous answer, "Robert Harvey".

<question> Who is the only other active Melbourne Demons player to have played in both the 1988 and 2000 AFL Grand Finals, alongside Steven Febey? </question>

<answer> Robert Harvey </answer>
===========================================================================

Now, answer this question again by yourself, pretending you do not know the ground-truth answer, the source document. You need to start answering this question from scratch. Most importantly, pretending you do not know the ground-truth answer, the source document, and do not mention the source document when thinking. If you find you lack some knowledge, you can call a search engine by <tool_call> query </tool_call> and it will return the top searched results between <tool_response> and </tool_response>. You can search as many times as your want. If you find no further external knowledge needed, you can directly provide the answer inside <answer> and </answer>, without detailed illustrations. For example, <answer> Beijing </answer>. Question: Who wrote a number of articles, profiles, and interviews of musicians between 2004 and 2006?

#### Assistant (turn 1)
<tool_call>\n{"name": "search", "arguments": {"query_list": ["1988 AFL Grand Finals players"]}}\n</tool_call>

<tool_response> ...... </tool_response>

#### Assistant (turn 2)
<tool_call>\n{"name": "search", "arguments": {"query_list": ["2000 AFL Grand Finals players"]}}\n</tool_call>

<tool_response> ...... </tool_response>

#### Assistant (turn 3)
I find that Robert Harvey and Steven Febey are the active Melbourne Demons players who have played in both the 1988 and 2000 AFL Grand Finals.
<answer> Robert Harvey </answer>

### Notice
Most importantly, pretending you do not know the ground-truth answer, the extra information in the source document, and do not mention the source document when thinking. 

###Begin
This is the reverse solution process  of the question (i.e., the process of generating the question) given by teacher:

===========================================================================
{demonstration}
===========================================================================

Now, answer this question again by yourself, pretending you do not know the ground-truth answer, the source document. You need to start answering this question from scratch. Most importantly, pretending you do not know the ground-truth answer, the source document, and do not mention the source document when thinking. If you find you lack some knowledge, you can call a search engine by <tool_call> query </tool_call> and it will return the top searched results between <tool_response> and </tool_response>. You can search as many times as your want. If you find no further external knowledge needed, you can directly provide the answer inside <answer> and </answer>, without detailed illustrations. For example, <answer> Beijing </answer>. Question: {question}
\end{lstlisting}
\end{tcolorbox}
\begin{tcolorbox}[
  enhanced,
  breakable,                 
  colback=white,
  colframe=black!50!white,
  title={User Prompt for the Teacher (Qwen3-4B / Qwen3-8B)},
  width=1.0\linewidth,        
  top=-5pt,        
  bottom=-5pt,     
  fonttitle=\small
]
\lstset{style=boxedcode}
\begin{lstlisting}[
  basicstyle=\ttfamily\scriptsize,
  breaklines=true,
  breakindent=0pt,
  breakautoindent=false,
  columns=fullflexible,
]
You are a helpful assistant. You will be given privileged information about the reverse solution process of the question (i.e., construction process of the question). Please pretend not to know the source document used to construct the question and the ground-truth in the reverse solution process, and answer the question again with reference to the construction process of the question.

###  General Rules
Precise Searching: When generating search queries, ensure they are specific, semantically complete, and directly target the key information you are missing.
Context Retention: Remember prior conversations and search results, maintaining logical consistency across multiple rounds of searching.
Termination Judgment: When information are sufficient to determine the answer, immediately stop searching and output the answer.
Reference to Privileged Information: When outputting search actions and answers, refer to the question construction process in privileged information, but do not directly output the ground-truth in the first turn or directly use extra information from the source document.
Search Boundary Constraint: When searching, do not introduce information that is not contained in the question or search results, even if it exists in the source document.

###  Example (Only showcase the logical style, please do not directly imitate the specific content)
#### User
This is the reverse solution process  of the question (i.e., the process of generating the question) given by teacher:
===========================================================================
Now, generate a question and its answer with n = 3 hops starting from the following source document: (Title: "Steven Febey")
selected in the national draft. He fell just short of Robert Flower's club record of 272 games and as of 2007 only four people have played more games for the Demons After the 2002 season, Febey went to Bali with his teammates and was caught up in the bombings which devastated the island. Standing beside former Demons player Steven Armstrong and David Robbins, Febey was about to enter the Sari Club when the car bomb exploded...[omitted due to page size]

Reasoning step 1: Identify Hop 1 in the document, formulate the query to reach Hop 2
<tool_call>\n{"name": "search", "arguments": {"query_list": ["other players who played in both 1988 and 2000 AFL grand finals for Melbourne Demons"]}}\n</tool_call><|im_end|>
.....

Hop 2 in the document and search results is not explicitly mentioned, so we need to further refine the search. The document mentions David Neitz, the captain of Melbourne, and players like Shane Woewodin and Greg Healy. However, the search results point us to the 1988 and 2000 AFL Grand Finals, indicating that we need to look for a player specifically from these finals who also played for Melbourne. The final hop, or Hop 3, must be a well-known player from the Melbourne Demons who played in both 1988 and 2000 grand finals, which can be Robert Harvey based on historical knowledge.
Reasoning step 2: Identify Hop 2 in the search results and reformulate to find Hop 3
<tool_call>\n{"name": "search", "arguments": {"query_list": ["Robert Harvey Melbourne Demons 1988 2000 grand finals"]}}\n</tool_call><|im_end|>
.....

The search results confirm that Robert Harvey played for the Melbourne Demons in both 1988 and 2000, making him the well-known player who fits the criteria. Therefore, Hop 3 is "Robert Harvey". The question is formulated to start with the explicit entity in the document, "Steven Febey", and lead to the single, unambiguous answer, "Robert Harvey".

<question> Who is the only other active Melbourne Demons player to have played in both the 1988 and 2000 AFL Grand Finals, alongside Steven Febey? </question>

<answer> Robert Harvey </answer>
===========================================================================

Now, answer this question again by yourself, pretending you do not know the ground-truth answer, the source document. 
You need to start answering this question from scratch. 
Most importantly, pretending you do not know the ground-truth answer, the source document, and do not mention the source document when thinking. 
You must conduct reasoning inside <think> and </think> first every time you get new information. 
After reasoning, If you find you lack some knowledge, you can call a search engine by <tool_call> query </tool_call> and it will return the top searched results between <tool_response> and </tool_response>. 
You can search as many times as your want. 
If you find no further external knowledge needed, you can directly provide the answer inside <answer> and </answer>, without detailed illustrations. 
Question: Who wrote a number of articles, profiles, and interviews of musicians between 2004 and 2006?

#### Assistant (turn 1)
<think>\nTo answer this question, I need to identify the players who played for the Melbourne Demons in both 1988 and 2000 AFL Grand Finals. The question mentions Steven Febey, and we need to find another player who played in both finals.
First, I will look for information about the 1988 and 2000 AFL Grand Finals and the players who participated in them. /
I will search for "1988 and 2000 AFL Grand Finals players" to find the relevant information.\n</think>

<tool_call\n{"name": "search", "arguments": {"query_list": ["1988 AFL Grand Finals players"]}}\n</tool_call>

<tool_response> ...... </tool_response>

#### Assistant (turn 2)
<think>\nI find the information about the 1988 AFL Grand Finals players. Next, I will search for "1988 and 2000 AFL Grand Finals players" to find the relevant information.\n</think>

<tool_call>\n{"name": "search", "arguments": {"query_list": ["2000 AFL Grand Finals players"]}}\n</tool_call>

<tool_response> ...... </tool_response>

#### Assistant (turn 3)
<think>\n  \n</think>
I find that Robert Harvey and Steven Febey are the active Melbourne Demons players who have played in both the 1988 and 2000 AFL Grand Finals.
<answer> Robert Harvey </answer>

### Notice
Most importantly, pretending you do not know the ground-truth answer, the extra information in the source document, and do not mention the source document when thinking. 

###Begin
This is the reverse solution process  of the question (i.e., the process of generating the question) given by teacher:

===========================================================================
{demonstration}
===========================================================================

Now, answer this question again by yourself, pretending you do not know the ground-truth answer, the source document. You need to start answering this question from scratch. Most importantly, pretending you do not know the ground-truth answer, the source document, and do not mention the source document when thinking. You must conduct reasoning inside <think> and </think> first every time you get new information. After reasoning, if you find you lack some knowledge, you can call a search engine by <tool_call> query </tool_call> and it will return the top searched results between <tool_response> and </tool_response>. You can search as many times as your want. If you find no further external knowledge needed, you can directly provide the answer inside <answer> and </answer>, without detailed illustrations. For example, <answer> Beijing </answer>. Question: {question}
\end{lstlisting}
\end{tcolorbox}
\captionsetup{justification=centering} 
\captionof{figure}{\textbf{Initial instructions for the teacher in $\pi$-play}}
\label{fig:user_prompt_teacher}
\end{center}

\subsection{Student Prompts}
\label{app:student_prompts}





\begin{center} 
\begin{tcolorbox}[
  enhanced,
  breakable,                 
  colback=white,
  colframe=black!50!white,
  title={User Prompt for the Student (Qwen3-4B-Instruct)},
  width=1.0\linewidth,        
  top=-5pt,        
  bottom=-5pt,     
  fonttitle=\small
]
\lstset{style=boxedcode}
\begin{lstlisting}[
  basicstyle=\ttfamily\scriptsize,
  breaklines=true,
  breakindent=0pt,
  breakautoindent=false,
  columns=fullflexible,
]
Answer the given question. If you find you lack some knowledge, you can call a search engine by <tool_call> query </tool_call> and it will return the top searched results between <tool_response> and </tool_response>. You can search as many times as your want. If you find no further external knowledge needed, you can directly provide the answer inside <answer> and </answer>, without detailed illustrations. For example, <answer> Beijing </answer>. Question: {question}
\end{lstlisting}
\end{tcolorbox}
\begin{tcolorbox}[
  enhanced,
  breakable,                 
  colback=white,
  colframe=black!50!white,
  title={User Prompt for the Student (Qwen3-4B / Qwen3-8B)},
  width=1.0\linewidth,        
  top=-5pt,        
  bottom=-5pt,     
  fonttitle=\small
]
\lstset{style=boxedcode}
\begin{lstlisting}[
  basicstyle=\ttfamily\scriptsize,
  breaklines=true,
  breakindent=0pt,
  breakautoindent=false,
  columns=fullflexible,
]
Answer the given question. You must conduct reasoning inside <think> and </think> first every time you get new information. After reasoning, if you find you lack some knowledge, you can call a search engine by <tool_call> query </tool_call> and it will return the top searched results between <tool_response> and </tool_response>. You can search as many times as your want. If you find no further external knowledge needed, you can directly provide the answer inside <answer> and </answer>, without detailed illustrations. For example, <answer> Beijing </answer>. Question: {question}
\end{lstlisting}
\end{tcolorbox}
\captionsetup{justification=centering} 
\captionof{figure}{\textbf{Initial instructions for the student in $\pi$-play}}
\label{fig:user_prompt_student}
\end{center}



\end{document}